\documentclass[a4paper,fleqn]{cas-dc}

\usepackage[authoryear]{natbib}

\def\tsc#1{\csdef{#1}{\textsc{\lowercase{#1}}\xspace}}
\tsc{WGM}
\tsc{QE}

\usepackage{graphicx}
\usepackage{amsmath}
\usepackage{amssymb}
\usepackage{booktabs}
\usepackage{multirow}
\usepackage{subcaption}
\usepackage{afterpage}
\usepackage[title]{appendix}
\hypersetup{
    breaklinks=true,
    colorlinks=true
}
\usepackage{xurl}

\begin{document}
\let\WriteBookmarks\relax
\def\floatpagepagefraction{1}
\def\textpagefraction{.001}

\shorttitle{Dense Temporal Contrast Synthesis via Conditioned Latent Transport}    

\shortauthors{S. Joshi et al.}  

\title [mode = title]{Dense Temporal Contrast Synthesis via Conditioned Latent Transport }  

\author[a]{Smriti Joshi}[orcid=0000-0001-8480-023X]
\cormark[1]
\ead{smriti.joshi@ub.edu} 
\credit{Conceptualization, Data curation, Methodology, Investigation, Project administration, Formal Analysis, Software, Writing - original draft, Writing - review \& editing}

\author[c]{Apostolia Tsirikoglou}\fnmark[$^\dagger$]
\credit{Data curation, Formal analysis, Writing - original draft, Writing - review \& editing}

\author[d,e]{Daniel M. Lang}\fnmark[$^\dagger$]
\credit{Formal analysis, Writing - review \& editing}

\author[a,g]{Richard Osuala}\fnmark[$^\dagger$]
\credit{Conceptualization, Writing - review \& editing}

\author[a,h]{Noah Márquez Vara}
\credit{Software}

\author[a]{Alejandro Guzman}
\credit{Data curation}

\author[a]{Grzegorz Skorupko}
\credit{Software}

\author[a]{Sebastian Ibarra Arregui}
\credit{Writing - review \& editing}

\author[a]{Lidia Garrucho}
\credit{Writing - review \& editing}

\author[i,j]{Akane Ohashi}
\credit{Data curation, Validation}

\author[k]{Dimitra Ntoula}
\credit{Data curation, Validation}

\author[l,m]{Eugen Divjak}
\credit{Validation, Writing - review \& editing}

\author[n]{O\u{g}uz Lafc\i}
\credit{Validation, Writing - review \& editing}

\author[g]{Jan C. Peeken}
\credit{Writing - review \& editing}

\author[d,e,f]{Julia A. Schnabel}
\credit{Writing - review \& editing}

\author[c]{Fredrik Strand}
\credit{Resources, Writing - review \& editing}

\author[a]{Oliver Diaz}
\credit{Supervision, Writing - review \& editing}

\author[a,b]{Karim Lekadir}
\credit{Supervision, Resources, Funding acquisition, Writing - review \& editing}
\cortext[1]{Corresponding author}


\affiliation[a]{organization={Departament de Matemàtiques i Informàtica, Universitat de Barcelona},
            city={Barcelona},
            country={Spain}}

\affiliation[b]{organization={Institució Catalana de Recerca i Estudis Avançats (ICREA)},
            city={Barcelona},
            country={Spain}}

\affiliation[c]{organization={Department of Oncology-Pathology, Karolinska Institutet},
            postcode={SE-171 77},
            city={Stockholm},
            country={Sweden}}

\affiliation[d]{organization={Institute of Machine Learning in Biomedical Imaging, Helmholtz Munich},
            city={Munich},
            country={Germany}}

\affiliation[e]{organization={School of Computation, Information and Technology, Technical University of Munich},
            city={Munich},
            country={Germany}}

\affiliation[f]{organization={School of Biomedical Engineering and Imaging Sciences, King's College London},
            city={London},
            country={United Kingdom}}

\affiliation[g]{organization={Department of Radiation Oncology, TUM University Hospital Rechts der Isar, TUM School of Medicine and Health, Technical University of Munich},
            city={Munich},
            country={Germany}}

\affiliation[h]{organization={Department of Computer Science and Engineering, Chalmers University of Technology},
            city={Gothenburg},
            country={Sweden}}

\affiliation[i]{organization={Department of Translational Medicine, Diagnostic Radiology, \& CIRCE -- the Center for Interdisciplinary Research on Cancer and Equity in Women, Lund University},
            city={Lund},
            country={Sweden}}

\affiliation[j]{organization={Department of Imaging and Physiology, Skåne University Hospital},
            city={Malmö},
            country={Sweden}}

\affiliation[k]{organization={Department of Radiology, Karolinska University Hospital}, postcode={17177}, city={Stockholm}, country={Sweden}}

\affiliation[l]{organization={University of Zagreb, School of Medicine},
            city={Zagreb},
            country={Croatia}}

\affiliation[m]{organization={University Hospital Dubrava},
            city={Zagreb},
            country={Croatia}}

\affiliation[n]{organization={Department of Biomedical Imaging and Image-Guided Therapy, Medical University of Vienna},
            city={Vienna},
            country={Austria}}

\nonumnote{$^\dagger$These authors contributed equally to this work.}

\begin{abstract}
Dynamic contrast-enhanced magnetic resonance imaging (DCE-MRI) is essential for breast cancer management, but reliance on gadolinium-based contrast agents (GBCAs) restricts use in contraindicated populations, prolongs scan protocols, and presents environmental toxicity concerns. Contrast synthesis offers a non-invasive alternative; however, existing approaches struggle to balance spatial realism with temporal continuity, suffer from slow iterative sampling, underutilize structural priors, and lack clinical validation. We propose a novel conditioned latent transport framework that predicts contrast enhancement in a single forward pass. By anchoring the latent trajectory to the pre-contrast anatomy and applying continuous time conditioning, the model synthesizes patient-specific contrast evolution at any given acquisition time. The proposed approach outperforms baseline and the state-of-the-art models across spatial, perceptual, temporal, and distributional metrics. Evaluated on an independent external cohort, the method demonstrates robustness to domain shifts induced by scanner noise as well as differing acquisition protocol. Furthermore, our synthetic contrast enhancement significantly improved downstream tumor segmentation performance, yielding a 22.4\% relative increase in Dice coefficient (0.60 vs. 0.49 baseline pre-contrast, $p < 0.01$), reducing boundary segmentation error by over 39\%, while outperforming all other generative model baselines. Finally, a reader study involving four breast radiologists evaluated the image quality, kinetic fidelity, and diagnostic viability of our synthesized sequences across 40 randomly selected cases. The results demonstrated that in 70\% of cases, synthesized images provided sufficient clinical information to support the same management decisions as real DCE-MRI, suggesting a path toward safer and faster contrast-free or contrast-reduced imaging workflows.
\end{abstract}


\maketitle
\section{Introduction}
\label{sec:intro}

Magnetic resonance imaging (MRI) is extensively used in clinical practice due to its high diagnostic sensitivity and soft-tissue contrast. It plays an integral role in breast cancer staging~\citep{staging1}, treatment monitoring, and surgical planning~\citep{surgical1, surgical2}. Unlike other imaging modalities like mammography and ultrasound, which are known to underestimate tumor extent~\citep{tumor_extent_anti_US_Mammo, tumor_extent_pro_MRI}, MRI enables the accurate delineation of tumor margins, potentially reducing surgical re-excision rates~\citep{reduced_operation}. 
In addition to treatment management, breast MRI is a critical screening modality for high-risk populations, including women carrying genetic mutations, those with a history of chest irradiation before age 30, or those presenting dense breast tissue~\citep{high_risk_1, high_risk_2}. The clinical value of MRI in screening is also significant; in one study, single-screening MRI has been shown to depict 18.1 additional cancers per 1,000 women with a history of breast-conserving therapy (BCT)~\citep{high_risk_better_sensitivity}. Moreover, the evaluation of time-intensity curves detailing contrast enhancement serves as a versatile biomarker, providing, among others, a reliable, non-invasive estimation of tumor malignancy~\citep{time_intensity_curve_1, time_intensity_curve_2}.

\begin{figure}
    \centering
    \includegraphics[width=0.9\linewidth]{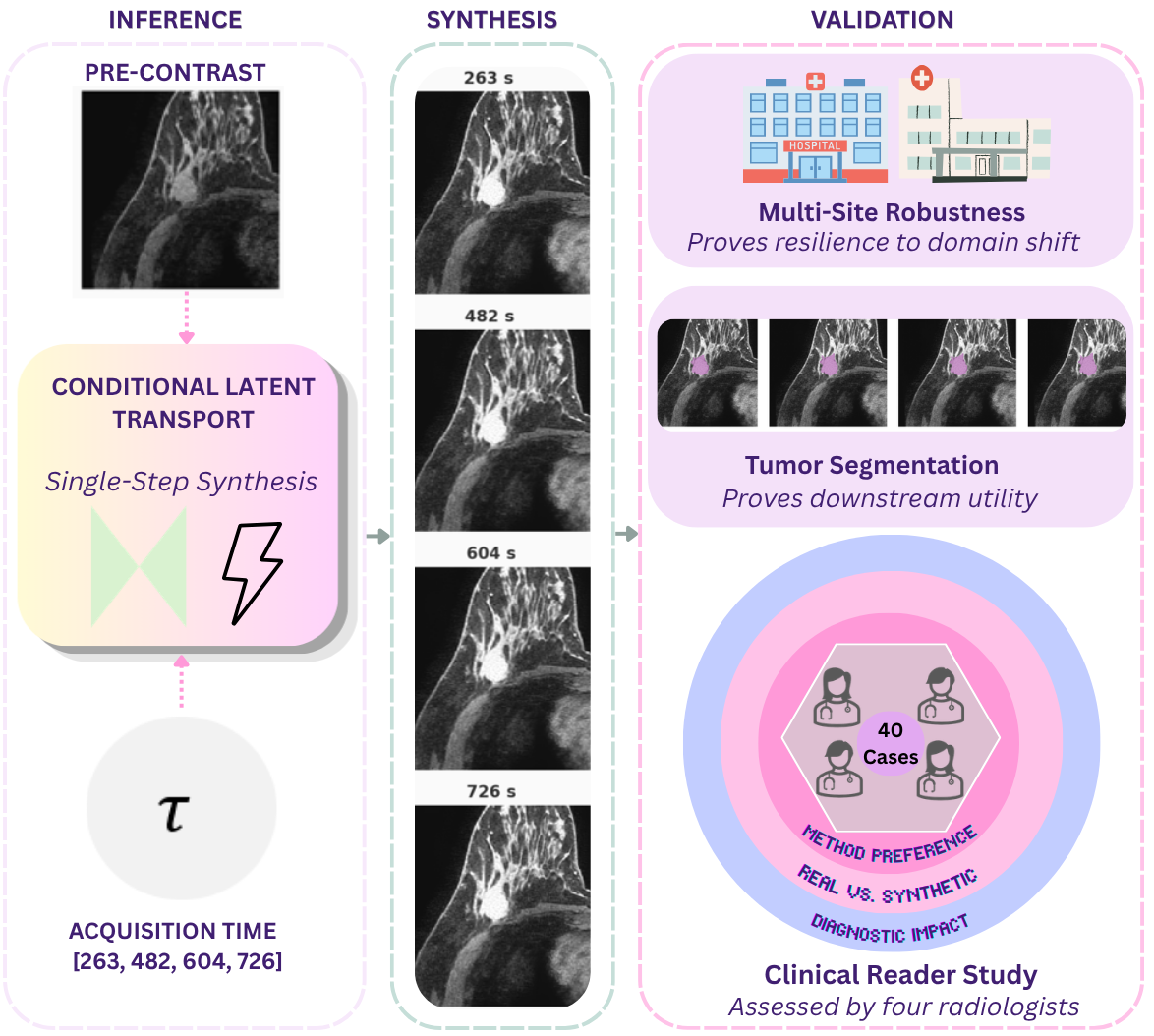}
    \caption{\textbf{Overview of the study design with validation framework.} Our proposed (left) method leverages a pre-contrast anchor and continuous acquisition time ($\tau$) to synthesize temporally consistent post-contrast DCE-MRIs (center). Clinical viability is evaluated via a three part validation strategy (right) comprising: (1) multi-site evaluation for domain robustness, (2) downstream tumor segmentation for biological fidelity, and (3) a clinical reader study to assess diagnostic impact.}
    \label{fig:paper_summary}
\end{figure}


The high sensitivity of MRI is largely owed to contrast agents~\citep{contrast_agents_breast_MRI}, most commonly gadolinium-based contrast agents (GBCAs). However, routine reliance on GBCAs introduces universal procedural burdens, alongside well-recognized toxicity risks~\citep{gadolinium_toxicity, gadolinium_risks, european_guidelines}. From an operational perspective, contrast administration requires intravenous administration, which invariably lengthens scan preparation time and carries a risk of access failure requiring specialized staff assistance. Furthermore, safety protocols mandate the immediate physical availability of a supervising physician, imposing geographical and scheduling restrictions on where and when breast MRI can be performed. In terms of patient safety, GBCAs pose significant risks and contraindications for vulnerable populations, including pregnant women~\citep{gd_pregnancy} and patients with renal failure~\citep{gd_renal}. They have also been associated with retention and accumulation in the brain~\citep{gd_brain} and bone tissue~\citep{gd_bone}, particularly following exposure to linear agents. Although newer macrocyclic agents appear to reduce tissue retention, they may not completely eliminate it~\citep{gd_macro}, and the clinical significance of this persistence remains uncertain. 

The environmental consequences of GBCAs are also considerable. Because these agents are predominantly excreted renally without undergoing metabolization, they resist degradation in wastewater treatment plants and are continuously emitted into aquatic ecosystems~\citep{gd_aquatic_1, gd_aquatic_2}. Ecologically, the $Gd^{3+}$ ion is capable of mimicking essential cations, including calcium, zinc, magnesium, and iron~\citep{gd_chemistry}, thereby threatening to disrupt critical cellular and biochemical pathways~\citep{gd_fish}. The potential for these agents to impact the food chain~\citep{gd_food_chain} through plants~\citep{gd_plants}, terrestrial as well as aquatic life~\citep{gd_terrestrial_aquatic}, alongside their unknown long-term ecological effects, constitutes a pressing and unresolved concern.

To mitigate these risks and reduce overall scan times, research has focused on synthesizing dynamic contrast-enhanced (DCE) MRI from non-contrast, early-phase, or low-dose acquisitions. Explored methods include generative adversarial networks (GANs)~\citep{dar2019image, muller2023using, osuala2025simulating, fonnegra2025synthesizing}, probabilistic diffusion models~\citep{ccnet, kishore2024dce, ibarra2025comparing, fan2025pre, kong2026mri}, and deterministic approaches~\citep{tenca, chung2025synthesizing, flowmi}. However, three primary limitations persist across these paradigms. First, enforcing temporal consistency across DCE sequences often compromises spatial realism. Realistic high-frequency textures are typically generated via stochastic mechanisms, such as noise injection or adversarial sampling; however, this inherent randomness introduces non-physiologically-grounded variations across sequential frames, disrupting smooth temporal continuity. Second, iterative generative models require extensive inference time, limiting computational efficiency. Third, existing methods underutilize pre-contrast anatomical priors; requiring a network to synthesize baseline structures from scratch distracts it from accurately modeling true contrast dynamics. Beyond these technical challenges, the true clinical utility of existing models remains difficult to assess due to a widespread lack of downstream task validation and expert reader evaluation. 

To address these gaps collectively, we propose a framework for dense temporal contrast synthesis, and rigorously evaluate it to position the method in terms of generalizability and clinical utility. A respective overview of this work is presented in Figure \ref{fig:paper_summary}. Our principal contributions are summarized as follows:

\begin{itemize}
   \item We introduce a novel conditioned latent transport network that synthesizes patient-specific DCE-MRI at any continuous acquisition time ($\tau$) from its corresponding pre-contrast image. 
   
   \item We demonstrate that anchoring the generative trajectory to the pre-contrast anatomy enforces macroscopic structural fidelity, while tailored spectral and structural loss functions preserve high-frequency details and perceptual realism.

    \item We evaluate our method on an independent, \textit{external validation cohort}, showing improvement over the pre-contrast image and competing state-of-the-art methods, while also systematically analyzing the domain shift and its effect on performance.

    \item We validate the anatomical fidelity of synthesized images through \textit{downstream tumor segmentation,} showing statistically significant improvements over competing methods across two complementary evaluation paradigms.

    \item We conduct a comprehensive \textit{clinical reader study} with four expert radiologists to assess the synthesized images, demonstrating their diagnostic viability and potential to reduce reliance on contrast administration in clinical practice.
\end{itemize}

The remainder of this manuscript is organized as follows: Section \ref{sec:related} reviews the current state-of-the-art in contrast synthesis. Section \ref{sec:method} outlines our proposed methodology. Section \ref{sec:experiments} discusses the implementation details. Section \ref{sec:results} discusses the quantitative and qualitative results. Finally, Section \ref{sec:conclusion} summarizes our core findings and highlights potential directions for future research.
\begin{figure*}[ht]
    \centering
    \includegraphics[width=\linewidth]{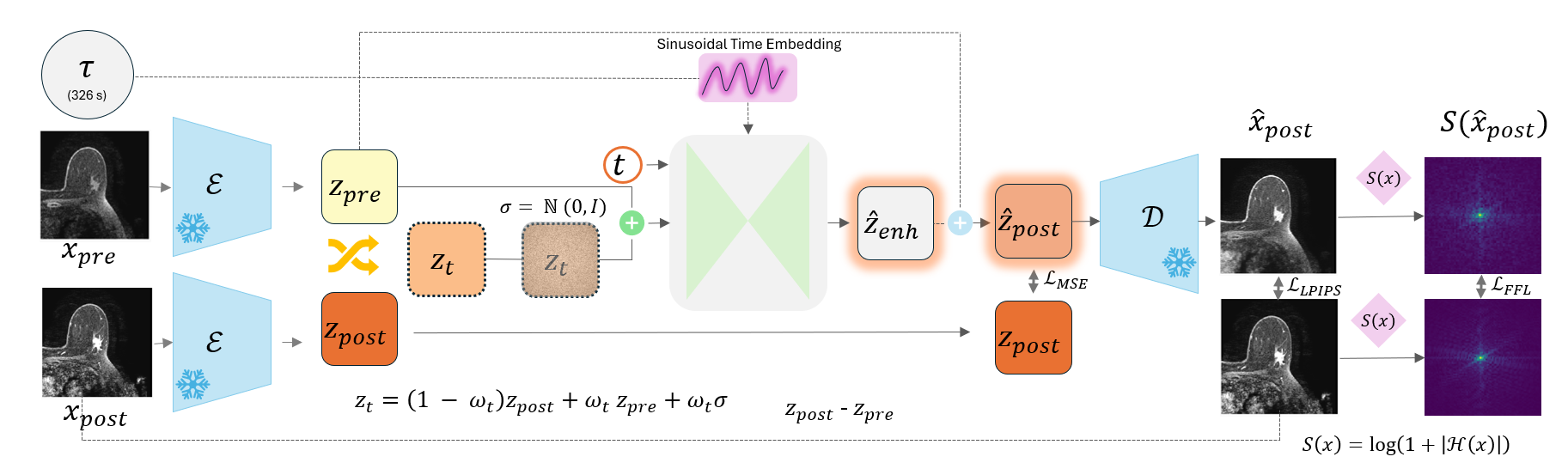}
    \caption{ Overview of the proposed Latent Generative Model Architecture. Pre-contrast ($x_{pre}$) and post-contrast ($x_{post}$) images are compressed into a high-fidelity latent space via a frozen encoder ($\mathcal{E}$) of a custom pretrained autoencoder. Then, the intermediate state ($z_t$) is constructed by interpolating between the non-enhanced structural anchor ($z_{pre}$) and the target state ($z_{post}$) with added stochasticity ($\sigma$). Conditioned on $z_{pre}$, a sinusoidal pharmacokinetic time embedding ($\tau$), and interpolation timestep t, the central U-Net acts as a target-predictor, directly regressing the constant latent subtraction map ($\hat{z}_{enh} \approx z_{post} - z_{pre}$). Adding the $z_{pre}$ to the predicted enhancement, $\hat{z}_{post}$ is obtained to compute the latent supervision loss ($\mathcal{L}_{MSE}$). Then, $\hat{z}_{post}$ is passed through a frozen decoder ($\mathcal{D}$) to upsample the image to pixel space. Structural and perceptual fidelity are strictly enforced via perceptual loss ($\mathcal{L}_{LPIPS}$), while a log-amplitude Fourier loss ($\mathcal{L}_{FFL}$) explicitly penalizes frequency deviations to preserve fine micro-vascular details.}
    \label{fig:network_architecture}
\end{figure*}
\section{Related Work}
\label{sec:related}

Early methodologies in contrast phase synthesis leveraged GAN based architectures such as pix2pix~\citep{isola2017image}, for image-to-image generation of post-contrast phases from pre-contrast sequences~\citep{osuala2025simulating, muller2023using}. GANs have also been applied to cross-phase synthesis, demonstrating the feasibility of predicting late post-contrast phases from early-phase acquisitions~\citep{fonnegra2025synthesizing, muller2024diffusion}. While these models successfully capture high-frequency structural details, they often struggle with training instability and the hallucination of non-existent anatomical features. 

Diffusion models~\citep{ho2020denoising, rombach2022high} provide an alternative to GANs because of their stable training dynamics and high-fidelity sample generation. In the medical imaging domain, diffusion models have been widely adopted across a spectrum of downstream tasks, including image synthesis~\citep{konz2024anatomically, pinaya2022brain}, artifact denoising~\citep{gao2023corediff}, and semantic segmentation~\citep{yan2024cold}.
Specifically, within contrast synthesis, ContrastControlNet (CCNet)~\citep{ccnet} trains the latent diffusion model for pre to post contrast synthesis in breast imaging, while parallel frameworks have been optimized for prostate imaging~\citep{prostate}. To further constrain the generative process and ensure anatomical fidelity, several approaches have incorporated multi-modal conditioning inputs, integrating raw imaging data with clinical metadata and explicit outline guidance~\citep{ibarra2025comparing,konz2024anatomically,fan2025pre,ccnet}.

Probabilistic diffusion models are valuable for modeling data diversity; however, the preference for consistent, repeatable outputs in clinical workflows motivates the shift toward deterministic synthesis. In this realm, works on breast DCE-MRI contrast synthesis rely on hierarchical networks~\citep{hierarchical}, temporal neural cellular automata~\citep{tenca}, and iterative networks~\citep{iterative}. These works promise a strong direction by demonstrating improved anatomical and temporal fidelity, which are frequently sacrificed in favor of perceptual realism within traditional stochastic models. Another line of research are cold diffusion models, which replace stochastic Gaussian perturbations with deterministic degradation operators (e.g. blurring) and learn the corresponding inverse restoration process~\citep{bansal2023cold}. Although not yet explored for contrast synthesis, this formulation has demonstrated promising results across a variety of medical imaging applications, including segmentation~\citep{cold_seg_1,cold_seg_2,cold_seg_3}, anomaly detection~\citep{cold_anomaly}, denoising~\citep{cold_noise_1,cold_noise_2}, and image reconstruction~\citep{cold_recon_1,cold_recon_2}. Importantly, the deterministic degradation trajectory provides a more structured restoration process compared to stochastic sampling, which can facilitate efficient inference. A parallel direction in generative modeling is represented by flow-based approaches, including flow matching and rectified flow formulations, which learn a time-dependent velocity field to transport samples between source and target distributions~\citep{lipman2022flow, liu2022flow}. These methods have been explored in latent generative frameworks~\citep{dao2023flow}, Computed Tomography (CT) image synthesis~\citep{wang2025ctflow}, multimodal MRI translation~\citep{tur2026wfm}, and more recently for contrast synthesis~\citep{flowmi}, where the task is formulated as a missing modality prediction problem.

Despite their impressive generative capabilities, both stochastic diffusion models and flow-based generative methods typically rely on multi-step denoising or numerical integration during inference. While effective for unconstrained image synthesis, repeated integration is not inherently required for contrast enhancement, where the source anatomy is already observed and only the physiological contrast uptake must be modeled, an observation also explored in conditional flow formulations~\citep{tur2026wfm}. Cold Diffusion replaces iterative denoising targets with direct endpoint prediction from deterministically corrupted intermediate states~\citep{bansal2023cold}. Concurrently, Rectified Flow frames generation as a transport problem learned from interpolated intermediate states, providing supervision along the transformation path rather than only at the endpoints~\citep{lipman2022flow}. Motivated by these observations, we formulate contrast synthesis as a conditional latent transport problem that combines endpoint prediction with structured interpolation, enabling efficient single-step inference while retaining the richer supervision provided by intermediate-state training.

\section{Methodology}
\label{sec:method}

Our method is presented visually in Figure \ref{fig:network_architecture}. We detail the mathematical formulation, architectural pipeline, and inference strategy of the proposed method below.

\subsection{Problem Formulation} 
Let $\mathcal{X} \subset \mathbb{R}^{C \times H}$ denote the pixel space of 2D MR image slices, where $H$ and $W$ represent the spatial height and width, respectively. For a given patient, standard clinical protocols acquire a non-enhanced pre-contrast baseline slice $x_{pre} \in \mathcal{X}$ and a target contrast-enhanced slice $x_{post}(\tau_i) \in \mathcal{X}$ captured at a discrete, protocol-specific physical timepoint $\tau_i$. However, while real scanner acquisitions are fundamentally discrete and temporally sparse, physiological contrast enhancement is a continuously evolving process. Therefore, our objective is to learn a continuous generative transport mapping $\mathcal{G}_\theta : (x_{pre}, \tau) \mapsto \hat{x}_{post}(\tau)$ that accepts any arbitrary timepoint $\tau \in \mathbb{R}^{+}$. This formulation allows us to accurately synthesize the non-linear pharmacokinetic hemodynamics of localized tumor regions across the entire temporal domain, while strictly preserving the underlying patient-specific anatomical structure and global tissue topology.

\subsection{Latent Space Compression and Custom Autoencoder}
To mitigate the severe computational bottlenecks associated with high-resolution medical imaging, we map the pixel space $\mathcal{X}$ into a compressed, low-dimensional latent manifold $\mathcal{Z} \subset \mathbb{R}^{c \times \frac{H}{f} \times \frac{W}{f}}$ using a Variational Autoencoder (VAE), where $f$ denotes the spatial downsampling factor. Let $\mathcal{E}$ and $\mathcal{D}$ denote the encoder and decoder, such that $z=\mathcal{E}(x)$ and $x \approx \mathcal{D}(z)$. 

Standard stable diffusion VAEs \citep{rombach2022high} typically employ an $f=8$ spatial downsampling factor, achieved through three consecutive convolutional blocks with a stride of 2. While computationally efficient, we empirically observed that this aggressive compression discards the fine-grained micro-vasculature and parenchymal textures essential for clinical diagnostics (Appendix \ref{appendix:vae}). To address this spatial bottleneck, we train a custom AutoencoderKL\footnote{https://huggingface.co/docs/diffusers/en/api/models/autoencoderkl} architecture from scratch, strictly constrained to a milder $f=4$ spatial downsampling factor. We optimize this encoder-decoder pair on our pre-contrast and post-contrast sequences using the standard VAE objective:
$$ \mathcal{L}_{VAE} = \mathcal{L}_{recon}(x, \mathcal{D}(\mathcal{E}(x))) + \beta \mathcal{D}_{KL}(q_\phi(z|x) || p(z)) $$,
where $\mathcal{L}_{recon}$ is a composite reconstruction loss combining Mean Squared Error (MSE) for signal fidelity and  Learned Perceptual Image Patch Similarity (LPIPS)~\citep{lpips} for structural sharpness, and $\mathcal{D}_{KL}$ is the Kullback-Leibler divergence used to regularize the latent manifold~\citep{kingma2013auto} by forcing the learned posterior $q_\phi(z|x)$ to approximate a standard normal prior $p(z)$. We observe that this $4\times$ configuration balances computational efficiency with the retention of micro-scale structures required for accurate contrast synthesis. Once trained, $\mathcal{E}$ and $\mathcal{D}$ are frozen. The resulting latent representations ($z_{pre}$ and $z_{post}$) provide the mathematical domain for the contrast synthesis network.

\subsection{Forward Process and Temporal Conditioning}
The core objective is to transport the non-enhanced baseline latent $z_{pre}$ to the target enhanced state $z_{post}(\tau)$. We frame this generation as a conditioned latent process governed by a linear interpolant. The network, parameterized as an acquisition-time conditioned U-Net $\mathcal{F}_\theta$, is trained to map a corrupted intermediate state $z_t$ to the fully enhanced target state. The forward process corrupts the target state directly toward the patient's non-enhanced baseline $z_{pre}$:
$$z_t = (1 - w_t) z_{post} + w_t z_{pre} + (w_t\cdot\sigma_{noise})\epsilon$$,
where $w_t \in [0, 1]$ dictates the temporal degradation schedule, $t \in [0, 1000]$ is the timestep, and $\epsilon \sim \mathcal{N}(0, I)$ introduces stochasticity to prevent deterministic collapse. The degradation weight $w_t$ governs the structural denoising process and is strictly decoupled from the physical acquisition time $\tau$. The stochastic noise variance scales linearly with $w_t$, maximizing at the baseline ($w_t=1$) and decaying to zero at the target ($w_t=0$).

We treat the physical acquisition time $\tau$ as a continuous conditional prior. It is encoded via Sinusoidal Positional Embeddings~\citep{vaswani2017attention} and passed through a Multi-Layer Perceptron (MLP) to match the generative network's feature dimensionality. This condition, alongside the timestep $t$, is injected into the U-Net residual blocks via Adaptive Group Normalization (AdaGN)~\citep{dharivwal2021diffusion}. Conditioning explicitly on $\tau$ enables the model to learn spatially-varying intensity mappings, assigning distinct physiological kinetics to different biological structures. The spatial input is constructed via channel-wise concatenation, yielding $z_{in} = [z_t \parallel z_{pre}]$. The generative mapping is thus formalized as $\hat{z}_{post} = \mathcal{F}_\theta(z_{in}, t, \tau)$.

\subsection{Residual Parameterization and Training Objective}
Because the non-enhanced anatomical topology remains strictly static during the localized acquisition window, the pharmacokinetic transformation can be expressed in residual form as $\Delta z = z_{post} - z_{pre}$. To improve representational efficiency, we parameterize the latent U-Net $\mathcal{F}_\theta$ to predict this residual component directly, yielding $\Delta \hat{z} = \mathcal{F}_\theta(z_{in}, t, \tau)$. This residual formulation decoupling the representation of the transformation (via $z_t$) from the prediction objective (via $\Delta z$) prevents the network from allocating capacity to redundant anatomical reconstruction to simplify optimization. The final enhanced latent is reconstructed as $\hat{z}_{post} = z_{pre} + \Delta \hat{z}$. 

We train the model using a composite objective function to ensure signal fidelity, perceptual realism, and the retention of fine frequency details: 
$$ \mathcal{L}_{total} = \lambda_{MSE} \mathcal{L}_{MSE} + \lambda_{LPIPS} \mathcal{L}_{LPIPS} + \lambda_{FFL} \mathcal{L}_{FFL} $$

\textit{Signal Fidelity ($\mathcal{L}_{MSE}$):} We apply a standard MSE loss between the predicted and ground-truth latents, defined as $||\hat{z}_{post} - z_{post}||_2^2$, to guarantee macroscopic pharmacokinetic accuracy and correct global intensity scaling.
 
\textit{Perceptual Realism ($\mathcal{L}_{LPIPS}$):} To overcome the deterministic blurring inherent to purely pixel-wise objectives, we evaluate LPIPS \citep{lpips} loss between the decoded images $x_{post}$ and $\hat{x}_{post}$. By comparing deep feature activations, this term enforces high-level structural coherence and prevents regression-to-the-mean artifacts.

\textit{Spectral Fidelity ($\mathcal{L}_{FFL}$):} Standard spatial losses often fail to resolve chaotic, high-frequency micro-vasculature. We incorporate a Focal Frequency Loss (FFL) \citep{jiang2021focal} to explicitly penalize discrepancies in the Fourier domain. Let $\mathcal{H}$ denote the 2D orthogonal Fast Fourier Transform. We map the predicted and ground-truth latents through the frozen decoder and penalize the logarithmic difference of their amplitude spectra:
$$\mathcal{L}_{FFL} = \left\| \log(1 + |\mathcal{H}(\mathcal{D}(\hat{z}_{post}))|) - \log(1 + |\mathcal{H}(\mathcal{D}(z_{post}))|) \right\|_1$$
This logarithmic scaling ensures that the network is mathematically incentivized to recover peripheral high-frequency clinical textures rather than being overpowered by the zero-frequency (DC) component. 

The values of $\lambda_{MSE}$, $\lambda_{LPIPS}$, and $\lambda_{FFL}$ were empirically determined on the validation set as 1, 5, and 50, respectively. 

\subsection{One-Step Inference Strategy}
During inference, given a pre-contrast anchor $z_{pre}$ and a requested target temporal phase $\tau$, the model evaluates the state at the maximum degradation timestep $t_{max}$ (effectively resulting in pre-contrast sequence with added noise $\sigma$). To preserve the generative capacity and mimic realistic scanner textures while ensuring smooth temporal consistency across continuous samples of $\tau$, we inject a fixed, patient-level latent noise map $\epsilon_{fixed}$. The noisy input state is constructed as:
$$z_{input} = z_{pre} + (w_{t_{max}} \cdot \sigma_{noise}) \epsilon_{fixed}$$,
where $w_{t_{max}}$ represents the terminal degradation weight and $\sigma_{noise}$ controls the scale of the injected generative variance. 

The U-Net receives this noisy input state $z_{input}$, concatenated with the clean anatomical anchor $z_{pre}$, the timestep $t_{max}$, and the continuous temporal condition $\tau$. In a single predictive pass, the network outputs the predicted residual $\Delta\hat{z}$, reconstructing the fully enhanced target endpoint as $\hat{z}_{post} = z_{pre} + \Delta\hat{z}$. Finally, the synthesized latent representation is mapped back to the high-resolution pixel space via the frozen VAE decoder, yielding the synthetic image $\hat{x}_{post} = \mathcal{D}(\hat{z}_{post})$.

\section{Implementation Details}
\label{sec:experiments}

This section describes the datasets, evaluation metrics, and baseline methods. Additional information, including data preprocessing, and network hyperparameters, can be found in Appendix \ref{appendix:contrast_synthesis}.


\begin{figure*}[htbp]
    \centering
    \begin{subfigure}[b]{0.30\linewidth}
        \centering
        \includegraphics[width=\linewidth]{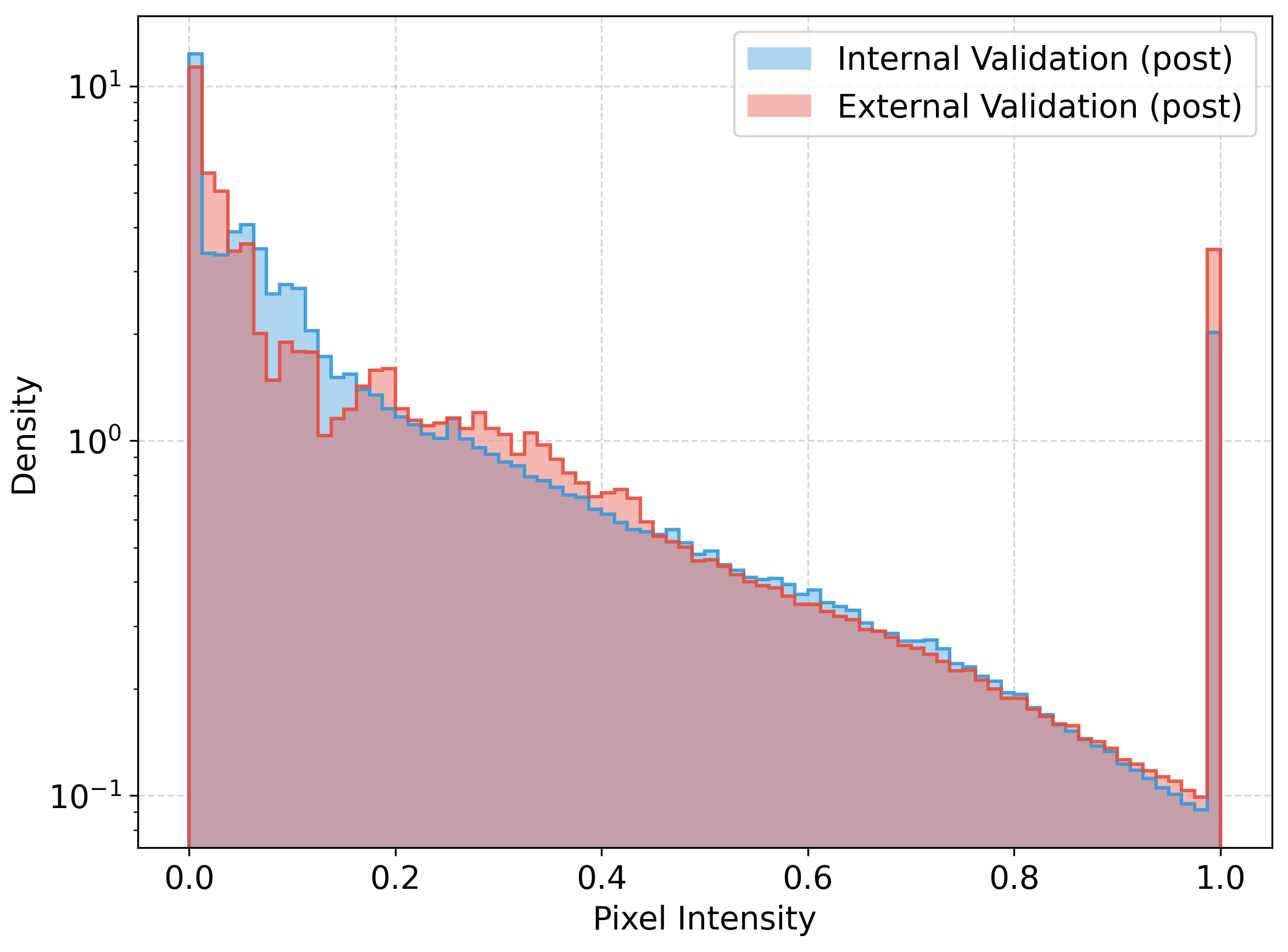}
        \caption{Intensity Histogram}
        \label{fig:intensity_histogram}
    \end{subfigure}
    \hfill
    \begin{subfigure}[b]{0.30\linewidth}
        \centering
        \includegraphics[width=\linewidth]{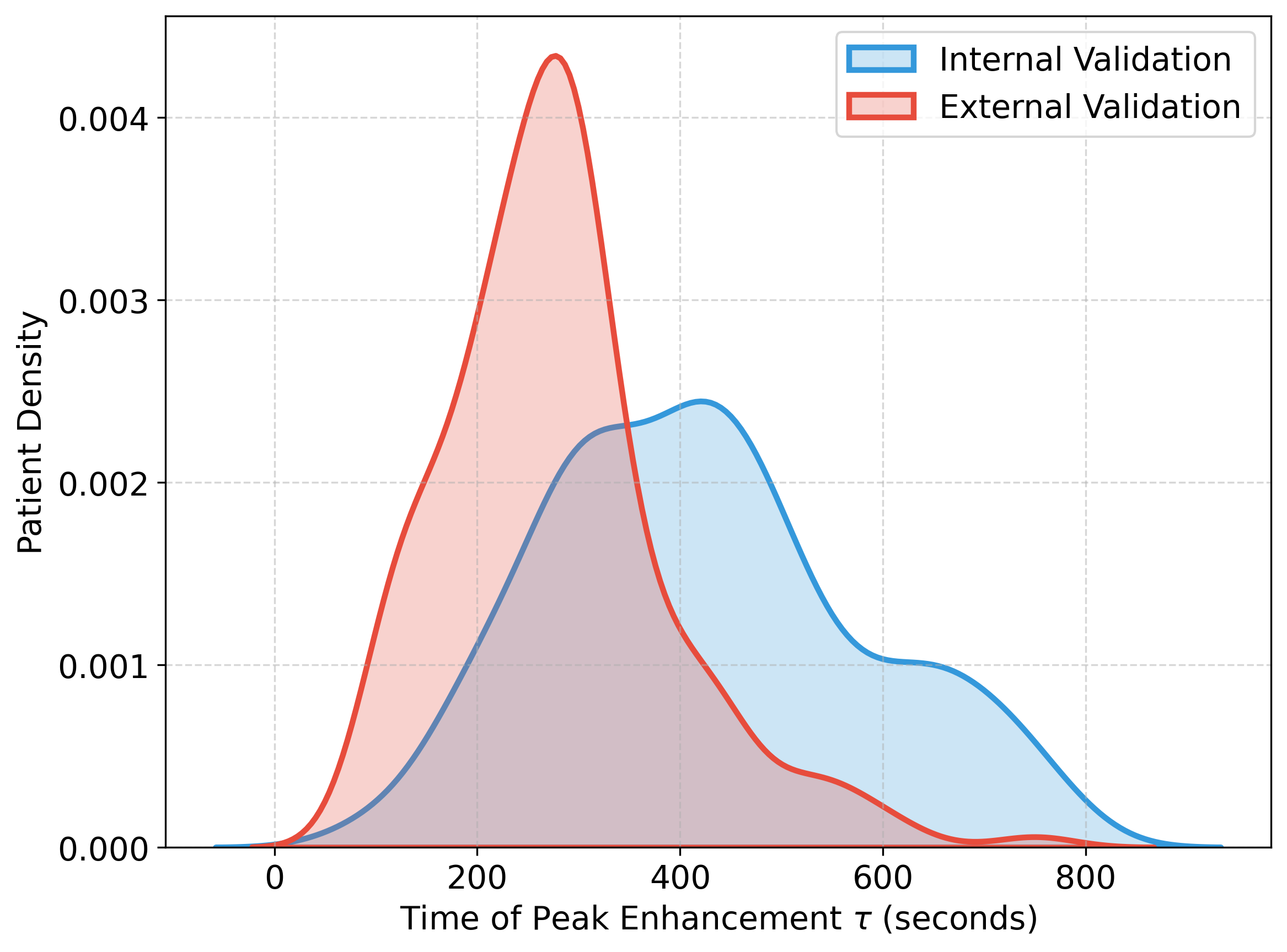}
        \caption{Peak Enhancement Time}
        \label{fig:temporal_gap}
    \end{subfigure}
    \hfill
    \begin{subfigure}[b]{0.28\linewidth}
        \centering
        \includegraphics[width=\linewidth]{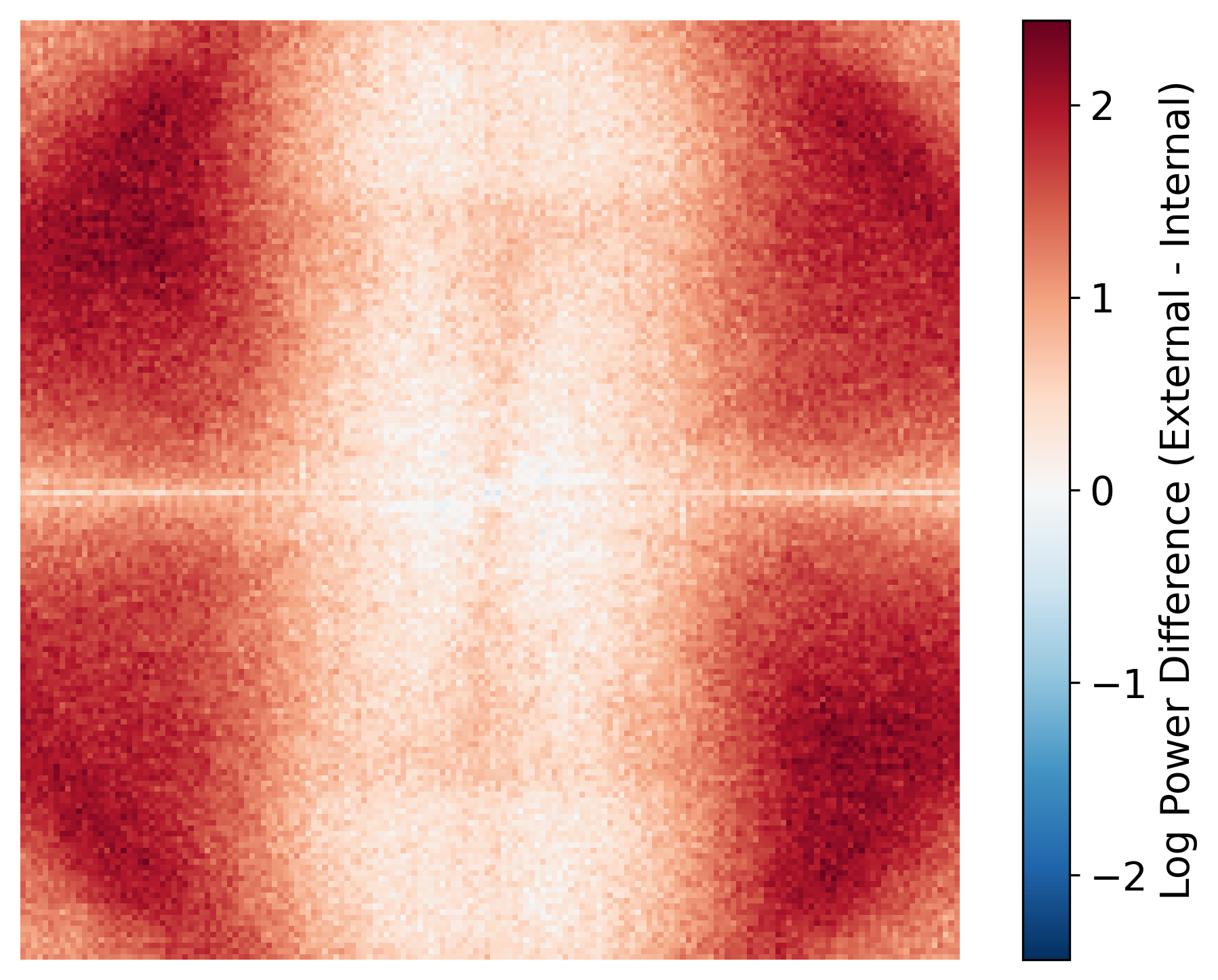}
        \caption{Spectral Difference}
        \label{fig:spectral_diff}
    \end{subfigure}
    
    \caption{Characterization of the domain shift between the internal and external validation cohorts. (a) Log-scaled marginal intensity histograms of the post-contrast phase exhibit general macroscopic alignment, indicating relatively comparable global contrast distributions. (b) However, the probability density function of peak enhancement time ($\tau$) reveals a severe temporal domain shift; the external cohort demonstrates significantly faster pharmacokinetic wash-in dynamics compared to the broader, delayed uptake of the internal dataset. (c) The 2D log power spectral difference between the cohorts highlights high-frequency spatial discrepancies, characteristic of differing scanner hardware, resolution, and acquisition protocols.}
    \label{fig:domain_gap}
\end{figure*}

\subsection{Datasets}
We use two publicly available datasets as well as one private dataset for training and validating our method.

\textbf{MAMA-MIA}: As detailed in ~\citep{garrucho2025large}, the MAMA-MIA dataset is a multi-center cohort of 1,506 patients featuring pre-treatment T1-weighted DCE breast MRIs. Compiled from the ISPY-1~\citep{ispy1}, ISPY-2~\citep{ispy2}, NACT~\citep{nact}, and Duke-Breast Cancer MRI~\citep{duke} collections, the dataset demonstrates high technical diversity, including various scanner vendors (GE, Siemens, Philips), magnetic field strengths (1.5T and 3T) and multiple acquisition planes (axial, sagittal). More importantly, MAMA-MIA provides expert tumor segmentations as well as harmonized acquisition times extracted in a uniform format from all the collections. Following the provided protocol, we maintain the designated split, utilizing 300 cases for the internal validation set.

\textbf{Duke-Breast-Cancer-MRI}~\citep{duke}: This dataset was collected from 2000-2014 and contains MRI scans of 922 patients acquired from two vendors, namely GE and Siemens, with 7 scanners of 1.5T and 3T magnetic field strengths from a single center in the United States. The acquisition plane is axial. MAMA-MIA includes 259 cases from this study. We incorporate the additional 663 cases from this collection in our training set.  

\textbf{Karolinska Instituet}: This private dataset was collected from 2012-2020 and contains MRI scans of 192 patients, acquired from a single vendor GE with 1.5T and 3T magnetic field strengths from a single center in Sweden. The acquisition plane is axial. We use this dataset to perform external validation and evaluate our model under domain shift.

\begin{table*}
\scriptsize
\centering
\caption{Quantitative evaluation of pharmacokinetic synthesis across pixel-level accuracy, structural fidelity, perceptual realism, and temporal alignment metrics. Baseline refers to the results on the pre-contrast image. MSE, DTW and DTW-ROI are reported in $10^{-2}$ scale. pix2pix model does not consider acquisition time $\tau$ during training. Therefore, temporal metrics are not reported for this method. Best results are highlighted in \textbf{bold}, second-best are \underline{underlined}. Statistical significance was evaluated using the Wilcoxon signed-rank test for all metrics, with the exception of PTE, which was assessed using a binomial sign test. Unless otherwise indicated, all reported values are statistically significant at $p < 0.001$. Markers denote lower significance levels: * indicates $p < 0.01$, ** indicates $p < 0.05$, and $\dagger$ indicates a statistically insignificant difference.}
\label{tab:indomain_results}
\begin{tabular}{lccccccccc}
\toprule
\rowcolor{pink!25}
\textbf{Method} & \textbf{MSE} $\downarrow$ & \textbf{PSNR} $\uparrow$ & \textbf{SSIM} $\uparrow$ & \textbf{LPIPS} $\downarrow$ & 
\textbf{FID} $\downarrow$ & \textbf{FRD} $\downarrow$ & \textbf{PTE} $\downarrow$ & \textbf{DTW} $\downarrow$ & \textbf{DTW-ROI} $\downarrow$\\
\midrule
\multicolumn{10}{l}{\textit{Internal Validation}} \\ \midrule

Baseline & 1.93 (1.39) & 18.24 (3.25) & 0.70 (0.11) & 0.19 (0.05) & 166.68 & 5.12 & 265.30 (114.43) & 2.91 (1.39) & 18.55 (5.46) \\
U-Net & 1.01 (0.67) & 20.84 (2.81) & 0.71 (0.11) & 0.19 (0.05) & 184.83 & 5.34 & \textbf{41.24} (83.58) $^\dagger$ & 0.84 (0.77) $^{**}$  & 6.18 (4.91) \\
pix2pix & 1.05 (0.60) & 20.36 (2.21) & \textbf{0.73} (0.08) & \textbf{0.18} (0.04) & 274.56 & \textbf{4.50} & -- & -- & -- \\
CCNet & 1.60 (0.87) & 18.60 (2.43)  & 0.58 (0.10) & 0.26 (0.05)  & 205.92 & 7.40 & 101.46 (108.67) & 1.21 (0.83) & 8.84 (0.05) \\
TeNCA & 0.99 (0.65) & 20.90 (2.69) & \textbf{0.73 (0.11)} & 0.21 (0.05) & 181.66 & 5.57 & 74.47 (110.09) $^{*}$ & 1.03 (0.86) & 5.97 (4.94) \\   
\textbf{Ours} & \textbf{0.84 (0.55)} & \textbf{21.61 (2.77)} & 0.71 (0.11) & \textbf{0.18} (0.05) & \textbf{154.56} & \underline{4.98} & \underline{44.57 (81.59)} & \textbf{0.68 (0.56)} & \textbf{3.76 (3.33)} \\\midrule

\multicolumn{10}{l}{\textit{External Validation}} \\ \midrule
Baseline & 2.52 (1.62) & 17.20 (3.73) & 0.69 (0.09)$^*$ & 0.20 (0.04) & \textbf{220.34} & 5.03 & 237.24 (104.89) & 3.31 (1.53) & 19.44 (4.54)  \\
U-Net & 1.26 (0.71) $^\dagger$ & 19.75 (2.71) & 0.70 (0.09) & 0.20 (0.04) & 257.74 & 4.93 & 43.59 (82.23)$^\dagger$ & 1.11 (0.84)$^\dagger$ & \underline{7.32 (4.67)$^\dagger$} \\
pix2pix & 1.22 (0.55) & 19.55 (1.90) & \textbf{0.73 (0.06)} & \textbf{0.19 (0.03)$^\dagger$} & 345.82 & \textbf{4.57} & -- & -- & --  \\
CCNet & 1.95 (1.01) & 17.68 (2.29) & 0.57 (0.09) & 0.26 (0.04) & 254.07  & 5.44 & 106.12 (95.11) & 1.27 (0.65) & 9.60 (4.60)  \\
TeNCA & \underline{1.11 (0.56)$^\dagger$} & 20.12 (2.40) & \underline{0.72 (0.08)} & 0.21 (0.04) & 244.23 & 5.22 & \underline{40.29 (77.88)$^\dagger$} & \underline{1.00 (0.62)$^\dagger$} & 8.25 (4.65)   \\
\textbf{Ours} & \textbf{1.08 (0.57)} & \textbf{20.32 (2.57)} & 0.69 (0.09) & \textbf{0.19 (0.04)} & \underline{226.26} & \underline{4.78} & \textbf{37.86 (70.13)} & \textbf{0.98 (0.68)} & \textbf{6.28 (3.95)}\\\bottomrule
\end{tabular}
\end{table*}

\begin{figure*}
    \centering
    \includegraphics[width=\linewidth]{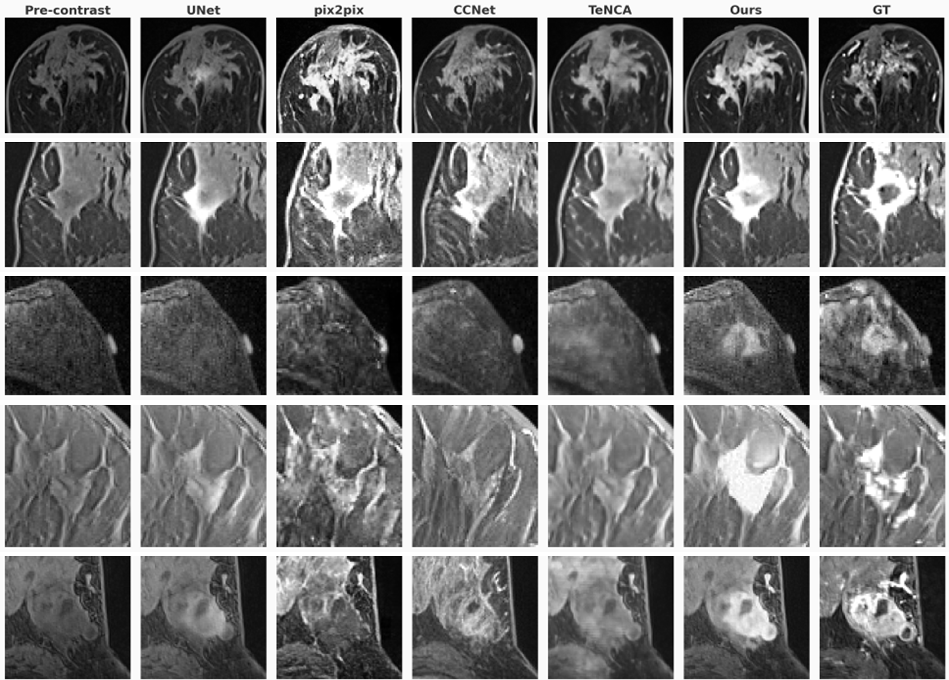}
    \caption{\textbf{Qualitative comparison of synthetic contrast generation.} This figure presents \textit{highly challenging cases} where tumor margins are unclear in the pre-contrast image, complicating accurate contrast injection for all models. The images displayed are selected and correspondingly synthesized at the earliest acquisition time available in the ground truth. Additional comparisons are provided in Appendix Figure \ref{appendix_fig:qualitative}. Columns from left to right: non-enhanced pre-contrast input, baseline methods (U-Net, pix2pix, CCNet, TeNCA), our method, and Ground Truth (GT). U-Net and TeNCA exhibit over-smoothing. pix2pix and CCNet demonstrate unpredictable behavior, capturing the structure in some instances but altering the underlying anatomy in others. Although our method also struggles to replicate exact tumor boundaries compared to the GT, it improves spatial fidelity and tumor localization.  Note that the figure presents magnified views centered directly on the tumor for better visibility, Images are normalized to [0, 255] for display.}
    \label{fig:qualitative}
\end{figure*}

\subsection{Metrics}

We evaluate the synthetic post-contrast MRIs using eight complementary metrics: MSE quantifies pixel-wise reconstruction fidelity. Peak Signal-to-Noise Ratio (PSNR) measures the overall reconstruction quality relative to the reference image. Structural Similarity Index Measure (SSIM) ~\citep{ssim} evaluates the preservation of local structural information, accounting for luminance, contrast, and texture consistency. LPIPS~\citep{lpips} measures perceptual similarity using deep feature representations. To assess distribution-level realism, we report DINOv2-based Fréchet Distance (FID-DINOv2)~\citep{fid, dinov2}. Specifically, we replace Inception features with self-supervised DINOv2 embeddings to better capture semantic and structural similarity in medical images. We also report the Fréchet Radiomics Distance (FRD)~\citep{frd}, computed from radiomic features extracted within the expert-annotated tumor masks. Unlike image-level metrics, FRD evaluates whether the generated images preserve clinically relevant tumor characteristics, including intensity statistics, texture, and shape-related descriptors, providing an assessment of radiomic fidelity. 

Furthermore, the quantitative evaluation of temporal dynamics in contrast synthesis remains largely absent from current literature. To address this gap and validate the temporal realism of the generated pharmacokinetic curves, we evaluate all methods using two targeted metrics:

\paragraph{Time-to-Peak Error (PTE):}

 In clinical BI-RADS assessment, the Time-to-Peak (TTP) is a critical diagnostic biomarker.

 To quantify temporal shift, we define the Peak Timing Error (PTE) as the mean absolute difference in seconds between the predicted and ground-truth time-to-peak across all temporal curves in the validation set. Let $T = \{\tau_1, \tau_2, \dots, \tau_K\}$ be the set of acquisition times, and $c(\tau)$ be the mean intensity of the tumor region at time $\tau$. The error is defined as:$$\text{TTPE} = \frac{1}{N} \sum_{i=1}^{N} \left| \underset{\tau \in T}{\arg\max} \; c_{pred}^{(i)}(\tau) - \underset{\tau \in T}{\arg\max} \; c_{gt}^{(i)}(\tau) \right|$$ A value of zero indicates that the model correctly identified the peak enhancement phase, while non-zero values correspond to a misalignment of one or more DCE phases (typically multiples of $\sim\!80$--$100$\,s, depending on the acquisition protocol).

\paragraph{Dynamic Time Warping (DTW):} 
To measure shape preservation independent of localized temporal distortions, we employ normalized Dynamic Time Warping (DTW)\citep{chung2025synthesizing} on the mean intensity value inside the ground truth mask. For a predicted sequence $P$ and ground truth sequence $G$, the optimal alignment path $\pi$ is computed to minimize the cumulative $L_1$ cost, normalized by the sequence lengths:$$\text{DTW}(P, G) = \frac{1}{|P| + |G|} \min_{\pi} \sum_{(i,j) \in \pi} | P_i - G_j |$$A lower DTW score indicates that the generative model successfully learned the continuous pharmacokinetic manifold, synthesizing biologically plausible enhancement trajectories even if subtly shifted in absolute time.

\subsection{Baseline Methods} To rigorously evaluate our proposed framework, we compare it against a diverse set of baseline and state-of-the-art techniques spanning multiple generative paradigms. First, we establish a standard U-Net~\citep{unet, schreiter2024virtual} which models sequential post-contrast sequences through different output channels. Second, we evaluate the adversarial pix2pix \citep{osuala2025simulating} architecture. Because this model does not incorporate continuous acquisition time conditioning, we exclude it from temporal metric evaluations and assess it solely on spatial image fidelity. Third, we compare against CCNet \citep{ccnet}, a latent diffusion model conditioned on both pre-contrast anatomy and acquisition time. Because CCNet synthesizes contrast by sampling from random noise, it exhibits high stochasticity and frame-to-frame variance. Finally, we benchmark against Temporal Neural Cellular Automata (TeNCA) \citep{tenca}, a deterministic approach based on neural cellular automata that generates smooth, dense temporal trajectories, originally designed to overcome the stochastic limitations of diffusion models like CCNet. U-Net and the proposed method rely on residual prediction, while the remainder of the methods directly predict post-contrast phases.

\section{Results}
\label{sec:results}

This section presents results on in-domain dataset, examines out-of-domain generalization, and ablates different components of the proposed method. The quantitative results are summarized in Table \ref{tab:indomain_results}.
\subsection{In-Domain Performance}

 At the pixel level, our method strictly dominates traditional reconstruction metrics, achieving the lowest MSE (0.84 x $ 10^{-2}$) and highest PSNR (21.61). While deterministic continuous-time models (TeNCA) and 2D adversarial networks (pix2pix) marginally outperform our framework on SSIM (0.73 vs. 0.71), this stems from inherent algorithmic biases: TeNCA favors smooth, regression-to-the-mean approximations, whereas pix2pix relies on hyper-realistic, albeit hallucinatory, textural synthesis (see Figure \ref{fig:qualitative}). This adversarial optimization also rewards pix2pix with lowest FRD of 4.50 on the test set, followed by our method at 4.98. Our formulation prioritizes true high-frequency structural coherence. This is quantitatively validated by our superior performance on LPIPS (0.18) and global feature distribution via FID-Dinov2 (154.56). Ultimately, the most significant advantage of our framework lies in its preservation of pharmacokinetic dynamics; our method drastically outperforms all baselines in temporal alignment, achieving a DTW of 0.68 (x $10^{-2}$) and a highly localized DTW-ROI of 3.76 (x $10^{-2}$). Our framework establishes a new state-of-the-art across the internal MAMA-MIA validation cohort, uniquely balancing spatial fidelity, perceptual realism, and temporal dynamics where existing baselines force a compromise.

Dense temporal contrast synthesis results can be viewed in \href{https://smriti-joshi.github.io/dense-temporal-contrast-synthesis/}{\textcolor{blue}{Supplementary Videos 1 - 3}}.

\subsection{Out-of-Domain Performance}

To fully contextualize the performance on the external KI cohort, we analyze the fundamental domain shifts between the two institutions in Figure \ref{fig:domain_gap}. First, the global post-contrast intensity distributions (Figure \ref{fig:intensity_histogram}) exhibit near-perfect alignment. This confirms the absence of macroscopic intensity covariate shifts; the physiological boundaries of contrast uptake remain strictly consistent across domains. This macroscopic alignment directly explains our model's robust preservation of structural realism (LPIPS: 0.19) and overall pixel accuracy.


However, density estimation of the acquisition priors (Figure \ref{fig:temporal_gap}) exposes a severe temporal concept shift. The external KI protocol captures peak enhancement significantly earlier ($\tau \approx 280$s) than the internal MAMA-MIA training distribution ($\tau \approx 450$s). While our temporal alignment metrics exhibit a predictable degradation compared to our internal validation baseline (DTW: 0.68 vs. 0.98), our method still achieves the best absolute temporal performance (Table \ref{tab:indomain_results}). Under this severe domain shift, the differences in global temporal alignment (PTE and DTW) between our approach, U-Net, and TeNCA do not reach statistical significance ($p > 0.05$). However, our framework maintains a statistically significant advantage over CCNet globally. Furthermore, within the localized tumor regions (DTW-ROI), our approach significantly outperforms both TeNCA and CCNet ($p < 0.001$), demonstrating robust preservation of tumor-specific kinetic fidelity despite the accelerated, out-of-distribution injection protocol.

Finally, while macroscopic statistics align, the 2D Spectral Difference Map (Figure \ref{fig:spectral_diff}) reveals a highly directional high-frequency covariate shift. The distinct spectral lobes indicate that the external scanner possesses a fundamentally different microscopic noise floor and k-space reconstruction profile. Therefore, the degradation observed in deep-feature metrics like FID (226.26) on the external cohort is likely an artifact of localized hardware noise and disparate institutional statistics, rather than a failure of the network's physiological synthesis.







\begin{table*}[t]
\scriptsize
\centering
\caption{\textbf{Ablation study.} Evaluating the impact of predicting subtraction, stochastic regularization and different loss components. Sub. refers to using subtraction between post-contrast and pre-contrast images as the prediction target, instead of the post-contrast image. $\epsilon$ refers to the added noise. $\mathcal{L}_{LPIPS}$ and $\mathcal{L}_{FFL}$ refer to the perceptual and spectral losses, respectively.  Best results are \textbf{bolded}.  MSE, DTW and DTW-ROI are reported in $10^{-2}$ scale.}
\label{tab:ablation}
\resizebox{\linewidth}{!}{
\begin{tabular}{cccc | ccccccccc}
\toprule
\rowcolor{pink!25}
\textbf{Sub.} & \textbf{$\epsilon$} & \textbf{$\mathcal{L}_{LPIPS}$} & \textbf{$\mathcal{L}_{FFL}$} &
\textbf{MSE} $\downarrow$ &
\textbf{PSNR} $\uparrow$ &
\textbf{SSIM} $\uparrow$ &
\textbf{LPIPS} $\downarrow$ &
\textbf{FID-Dinov2} $\downarrow$ &
\textbf{FRD} $\downarrow$ &
\textbf{PTE} $\downarrow$ &
\textbf{DTW} $\downarrow$ &
\textbf{DTW-ROI} $\downarrow$ \\
\midrule
 & & \checkmark && 
0.93 (0.53) & 20.98 (2.49) &\textbf{0.71 (0.11)} & \textbf{0.18 (0.05)} & 162.83 & 5.17 & 45.35 (77.37) & 0.94 (0.64) & \textbf{3.48 (3.47)} \\

 \checkmark & & & & 
1.06 (0.63) & 20.53 (2.65) & 0.68 (0.12) & 0.22 (0.06) & 357.52 & 7.81 & 48.94 (80.26) & 1.70 (1.07) & 3.54 (3.32) \\

\checkmark & & \checkmark & & 
 0.89 (0.54) & 21.31 (2.66) & \textbf{0.71 (0.11)} & \textbf{0.18 (0.05)} & 152.14 & 5.10 & 47.87 (78.60) & 0.80 (0.61) & 4.02 (3.61) \\


\checkmark & \checkmark & \checkmark && 
0.86 (0.54) & 21.47 (2.74) &\textbf{0.71 (0.11)} &
\textbf{0.18 (0.05)} &
\textbf{145.06} & 5.05 &
45.65 (83.22) &
0.77 (0.62) & 3.65 (3.23) \\

\checkmark & \checkmark & \checkmark & \checkmark & 
 \textbf{0.84 (0.55)} & \textbf{21.61 (2.77)} & \textbf{0.71 (0.11)} & \textbf{0.18 (0.05)} & 154.56 & \textbf{4.98} & \textbf{44.57 (81.59)} & \textbf{0.68 (0.56)} & 3.76 (3.33) \\
\bottomrule
\end{tabular}
}
\end{table*}

\begin{figure*}
    \centering
    \includegraphics[width=\linewidth]{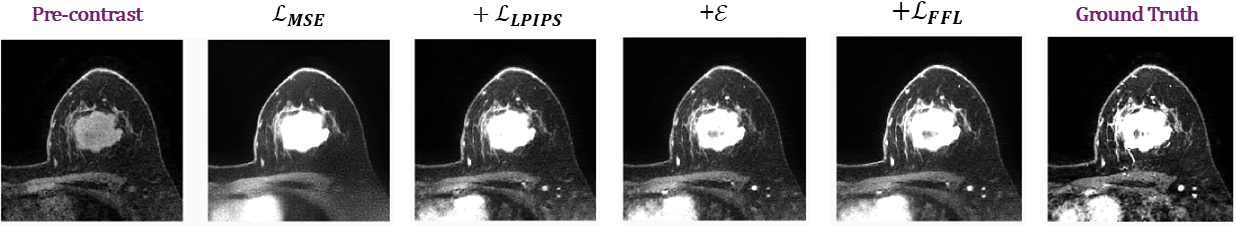}
    \caption{Qualitative evaluation of added ablation components. These images are normalized to range [0, 255] for display.}
    \label{fig:ablation}
\end{figure*}

\subsection{Ablation Study}

\subsubsection{The Impact of Loss Components} 

Our ablation study (Table \ref{tab:ablation}) and corresponding qualitative visualizations (Fig. \ref{fig:ablation}) demonstrate the critical, additive benefits of integrating structural, perceptual, and frequency-based constraints into the synthesis pipeline. Training with $\mathcal{L}_{MSE}$ maps the macro-level contrast uptake, but visually results in a highly over-smoothed, blurry lesion devoid of internal heterogeneity. Introducing perceptual loss mitigates this blurring, recovering structural dimension and sharpening macroscopic tumor boundaries. The subsequent addition of stochastic regularization (+ noise) injects essential micro-textural realism into the tissue, visually harmonizing the generated distribution and quantitatively improving deep-feature semantic fidelity (optimizing FID-Dinov2 to 145.06).

Finally, the integration of Fourier-domain guidance proves essential for recovering high-frequency morphological details. As observed visually, the Fourier constraint improves contrast in fine vascular structures surrounding the tumor area, closely aligning the generated output with the ground truth. Quantitatively, this comprehensive formulation yields the best overall pixel fidelity (MSE: 0.0084, PSNR: 21.61), radiomic texture preservation (FRD: 4.97), and temporal pharmacokinetic accuracy (DTW: 0.0068).

\subsubsection{Effect of pre-contrast concatenation}

\begin{table}[htbp]
\centering
\caption{\textbf{Ablation Study.} Effect of Pre-conditioning. Best results are highlighted in \textbf{bold}.  MSE, DTW and DTW-ROI are reported in $10^{-2}$ scale.}
\label{tab:ablation-preconditioning}
\begin{tabular}{lrr}
\toprule
\rowcolor{pink!25}
\textbf{Metric} & \textbf{w/o pre conditioning} & \textbf{w pre conditioning} \\
\midrule
\textbf{MSE $\downarrow$} & 1.04 (0.66) & \textbf{0.84 (0.55)} \\
\textbf{PSNR $\uparrow$} & 20.67 (2.73) & \textbf{21.61 (2.77)} \\
\textbf{SSIM $\uparrow$} & \textbf{0.71 (0.11)} & \textbf{0.71 (0.11)} \\
\textbf{LPIPS $\downarrow$} & 0.18 (0.05) & \textbf{0.18 (0.05)} \\
\textbf{PTE $\downarrow$} & 58.94 (92.79) & \textbf{44.57 (81.59)} \\
\textbf{FID-Dinov2 $\downarrow$} & \textbf{141.87} & 154.56 \\
\textbf{FRD $\downarrow$}  & 5.67 & \textbf{4.98} \\
\textbf{DTW $\downarrow$} & 1.08 (1.01) & \textbf{0.68 (0.56)} \\
\textbf{DTW-ROI $\downarrow$} & 4.45 (3.73) & \textbf{3.76 (3.33)} \\
\bottomrule
\end{tabular}
\end{table}

Table \ref{tab:ablation-preconditioning} evaluates the effect of the explicit morphological anchor ($z_{pre}$) on generative fidelity. The unanchored model achieves a lower FID-Dinov2 score (141.91 vs. 155.17) and a similar SSIM. By synthesizing both baseline anatomy and contrast through a single pathway, the unanchored network produces smooth deep-feature representations. However, explicitly conditioning the vector field on $z_{pre}$ improves the clinical and temporal metrics. The anchored network focuses on contrast synthesis, achieving better pixel-level calibration (MSE: 0.0084), radiomic texture preservation (FRD: 4.97), and reduced trajectory error (DTW: 0.0068 vs. 0.0098). 

\begin{figure}
    \centering
    \includegraphics[width=0.9\linewidth]{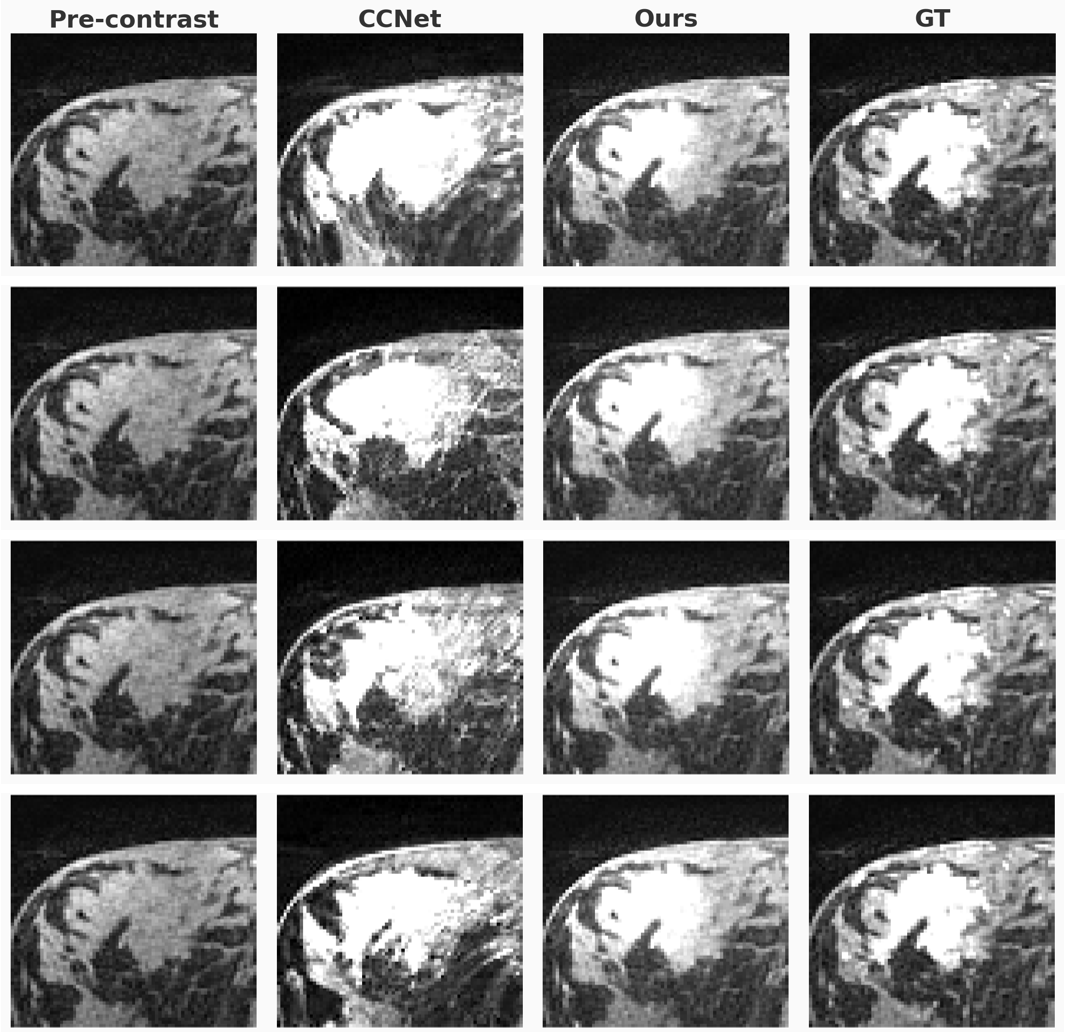}
    \caption{\textbf{Qualitative comparison of multiple stochastic predictions.} Columns refer to pre-contrast image, synthesized images from CCNet \citep{ccnet} and our method respectively, and GT (real post-contrast image). Four rows correspond to four different random noise initializations. In CCNet, the injected noise leads to structurally different anatomical predictions, including the shape of the tumor region. In the proposed method, anatomy is strictly preserved and noise accounts for subtle contrast variations and inherent MRI noise. Note that the figure presents magnified views centered directly on the tumor for better visibility, These images are normalized to range [0, 255] for display.}
    \label{fig:pre_contrast_noise}
\end{figure}

Furthermore, while both CCNet and our approach use stochastic noise to model output distributions, the pre-contrast anchor and the amount of injected noise dictate its specific function in image generation. CCNet initializes from pure noise conditioned on the pre-contrast image and acquisition time. As shown in Figure \ref{fig:pre_contrast_noise}, while this conditioning preserves the macroscopic structure of the breast, different random noise result in varying tumor extents and anatomical hallucinations, across four predictions at a given acquisition time. In contrast, our method uses the pre-contrast image as an anchor to retain the macroscopic integrity of both the anatomy and the pathology. Consequently, the injected noise models epistemic uncertainty, manifesting as realistic variations in contrast and standard MRI noise. The relevance of this property for uncertainty estimation is discussed in Section \ref{sec:reader_study}.

\subsubsection{How to add noise for temporal continuity?} Standard clinical practice typically acquires 4-5 post-contrast phases during MRI exam. Transitioning from this sparse sampling to modeling a dense temporal trajectory requires balancing spatial realism with temporal continuity. While injecting latent noise mimics uncertainty in contrast and scanner texture, independent random sampling across frames causes non-physiological flickering. We compared two noise injection strategies during inference, \textit{Independent random noise} per phase restores spatial texture but disrupts temporal continuity. \textit{Patient-level noise} holds a single noise map constant across all phases for a given patient, ensuring the temporal condition solely drives image changes. While we compare CCNet with independent noise to replicate the original setting of the paper, we also demonstrate that patient-level noise can improve the method to yield smooth trajectories. Specifically, this can be done because CCNet is based on denoising diffusion implicit models, which do not require noise injection at each denoising step. In \href{https://smriti-joshi.github.io/dense-temporal-contrast-synthesis/}{\textcolor{blue}{Supplementary Videos 4 - 7}}, we show the comparison for both CCNet and the proposed method with different noise strategies. We observe that patient-level noise results in smoother trajectories for both methods. In contrast to our method, which maintains reasonable consistency across different random noise seeds, CCNet exhibits a high degree of variance and becomes unreliable due to its inherent dependence on noise, reinforcing the observations made in the previous section.

\subsection{Downstream Tumor Segmentation}

\begin{table}
\centering
\caption{\textbf{Quantitative downstream segmentation performance on the MAMA-MIA internal validation set.} Results evaluate the performance on synthesized images against real pre-contrast (Baseline) and real post-contrast (Upper Bound) targets. Statistical significance was computed via a paired Wilcoxon signed-rank test ($p < 0.01$). Results that did not reach statistical significance are marked with an asterisk ($*$).}
\label{tab:segmentation}
\begin{tabular}{lrr}
\toprule
\rowcolor{pink!25}
\textbf{Method} &  \textbf{Dice Coefficient} $\uparrow$ & \textbf{HD95} $\downarrow$ \\
\midrule
\multicolumn{3}{@{}l}{\textit{Segmentation model trained on post-contrast}} \\ \midrule
Baseline   & 0.17 (0.30) & 160.69 (104.48) \\
U-Net                       & 0.44 (0.36) & 82.20 (104.68)* \\
pix2pix & 0.22 (0.33) & 147.79 (108.24) \\
CCNet          &  0.45 (0.35)           &  76.81 (102.32)               \\
TeNCA        & 0.30 (0.35) & 122.31 (110.69) \\
\textit{Ours}               & \textbf{0.51 (0.37)} & \textbf{68.78 (99.62)} \\
Upper Bound & 0.68 (0.33) & 38.59 (77.73)  \\

\midrule
\multicolumn{3}{@{}l}{\textit{Segmentation model trained on pre-contrast}} \\ \midrule
Baseline   & 0.49 (0.37) & 71.48 (99.90) \\
U-Net                       & 0.56 (0.35) & 53.05 (89.01) \\
pix2pix & 0.44 (0.35) & 70.10 (96.55)  \\
CCNet        &  0.44 (0.35) &  78.51 (102.96) \\
TeNCA          & 0.51 (0.36) & 62.59 (93.86) \\
\textit{Ours}               & \textbf{0.60 (0.33)} & \textbf{43.38 (80.12)} \\
Upper Bound & 0.63 (0.34) & 47.46 (84.74) \\
\bottomrule
\end{tabular}

\end{table}

\begin{figure*}
    \centering
    \includegraphics[width=\linewidth]{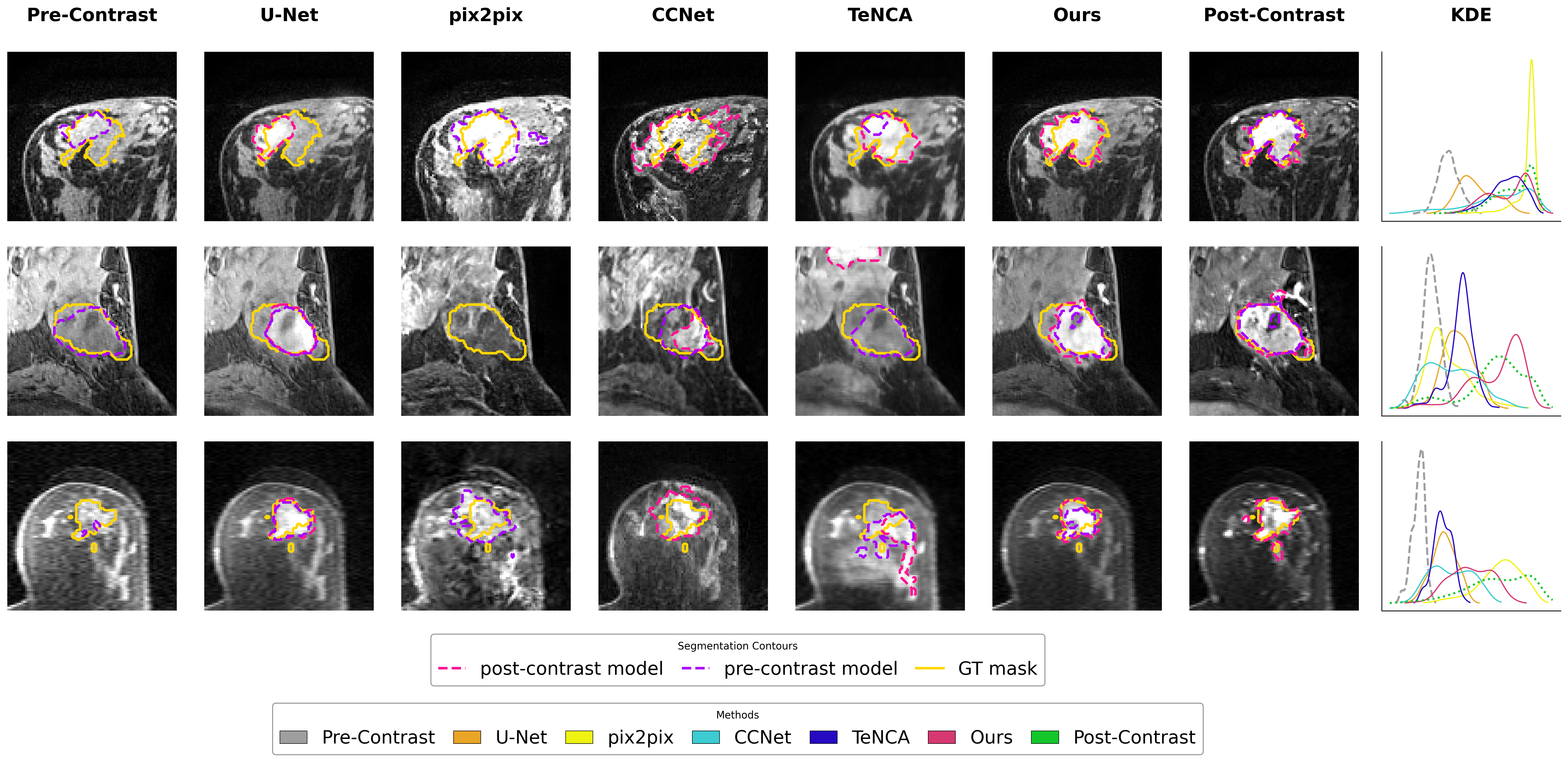}
    \captionof{figure}{\textbf{Qualitative downstream segmentation results.} Each row depicts a distinct patient case across real and synthesized images. Note that the figure presents magnified views centered directly on the tumor for better visibility, These images are re-normalized to range [0, 255] for display. The accompanying Kernel Density Estimation (KDE) maps (far right) illustrate the intensity distribution within each ground truth masks. The \textbf{\textit{top row}} features a \textit{large, highly lobulated malignant lesion} exhibiting limited signal on the pre-contrast scan. The pix2pix model artificially spikes pixel intensity (evidenced by the sharp green peak in the KDE plot) in and around the tumor, resulting in poor boundary discrimination. CCNet hallucinates high level of texture and overestimates the tumor region. While U-Net underestimates the tumor extent and TeNCA demonstrates moderate localization, our model achieves the closest alignment with the ground truth (GT) in both morphological boundary and intensity distribution.
    The \textbf{\textit{middle row}} presents a \textit{diffuse, lower-contrast malignant lesion}, visible merely as a darker hypointense region in the baseline scan but still, correctly identified by the pre-contrast trained model. Here, CCNet, and U-Net underestimates the region, pix2pix fails to provide meaningful enhancement, and TeNCA suffers from a spatial hallucination (false positive). Conversely, our model successfully recovers the broad enhancement profile, matching both the GT boundaries and the true KDE distribution.
    The \textbf{\textit{bottom row}} depicts a \textit{distinct malignant focal mass}. Although pix2pix achieves a KDE profile closer to the GT, it suffers from severe image degradation and, similar to TeNCA, massively overestimates the tumor boundary. While U-Net, CCNet  and our proposed method accurately localize the mass, our method demonstrates superior statistical consistency with the true post-contrast KDE profile.}
    \label{fig:segmentation_visuals}
\end{figure*}

To evaluate the preservation of biological structures, we compare downstream tumor segmentation performance on real and generated scans (Table \ref{tab:segmentation}). 

Specifically, two 2D nnUNet~\citep{isensee2021nnu} models are trained on the MAMA-MIA training set, on (a) first post-contrast images only, and (b) pre-contrast images only. We evaluate downstream segmentation using the Dice coefficient and the 95th percentile Hausdorff Distance (HD95). These metrics are reported exclusively for the first post-contrast phase, as the ground truth masks were delineated on this specific timepoint. Training details, extended evaluations across all subsequent temporal sequences, as well as comprehensive results on the external validation cohort, are detailed in the Appendix \ref{appendix:segmentation}.

\textit{Can a segmentation model trained on real post contrast images find the tumor on virtually contrast injected images?} 

On the pre-contrast baseline, the post-contrast trained segmentation network effectively fails to localize the lesion (Dice: 0.17), establishing its critical reliance on enhancement. While the competing methods including pix2pix and TeNCA previously demonstrated high SSIM, their corresponding downstream segmentation performances only show a small improvement over the non-enhanced baseline, In contrast, our proposed method explicitly preserves biological boundaries, achieving the highest segmentation performance (Dice: 0.51, HD95: 68.78) and recovering the vast majority of the true clinical signal (Upper Bound Dice: 0.67). 

\textit{If the goal is to avoid contrast, why not simply rely entirely on pre-contrast scans? In other words, why synthesize at all?} 

Segmentation networks trained directly on post-contrast data rely highly on the magnitude of pixel intensity \citep{joshi2024leveraging}, as evident by low performance on pre-contrast data. Consequently, they also have higher sensitivity to variations in contrast dynamics, for instance, degree of contrast uptake. Conversely, as shown in Table \ref{tab:segmentation}, a network explicitly trained on pre-contrast data learns robust morphological and textural priors and already localizes tumor better in absence of contrast signal (Dice: 0.49). When we route these images through the generative frameworks, synthetic contrast surfaces the latent physiological structures, and all deterministic methods improve over the baseline. Our method performs the best with Dice of 0.60 and an HD95 of 43.38 (Upper bound: 0.63), thereby maximizing the algorithmic utility of raw scans while bypassing the physiological risks of actual gadolinium.

Visual inspection (Figure \ref{fig:segmentation_visuals}) shows three representative examples where our model produces intensity distributions (pink) that closely mirror the true post-contrast profiles (green), and yields automated segmentation boundaries that conform to the ground truth. Additionally, our proposed framework exhibited the lowest absolute failure rate across all evaluated methods and training paradigms, yielding complete missed detections in only 29 of 300 cases. This performance was second only to the true post-contrast ground truth (16/300).  A detailed analysis of catastrophic segmentation failures (Dice = 0) is provided in the Appendix Section \ref{appendix:segmentation_failure} and qualitative examples are added to Appendix Figure \ref{appendix_fig:segmentation_failure}.

\begin{figure*}
    \centering
    \includegraphics[width=\linewidth]{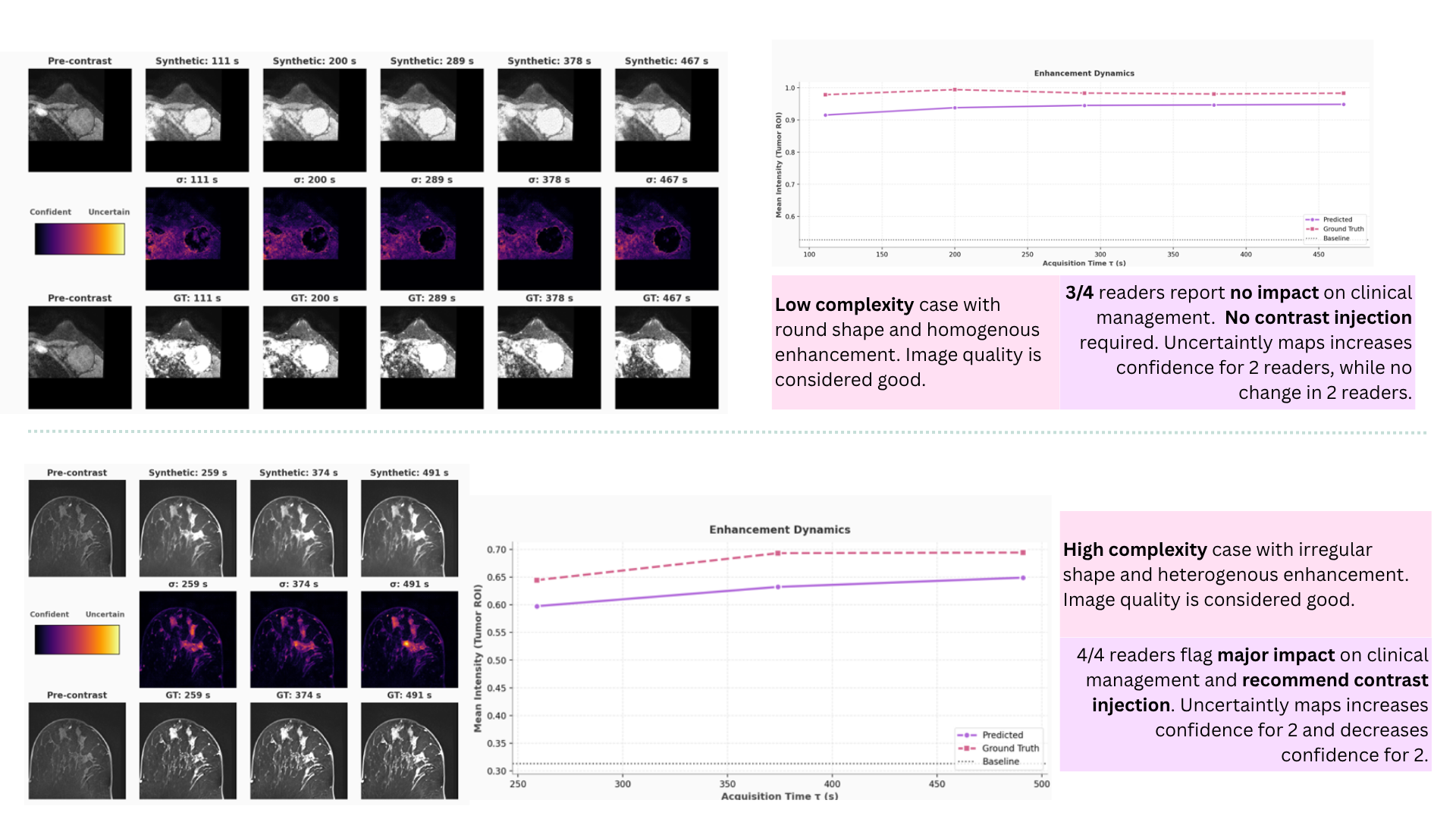}
    \caption{\textbf{Clinical evaluation across varying case complexities.} The figure contrasts a representative low-complexity case (Top) with a high-complexity case (Bottom) from the reader study. For both cases, the image panels display the pre-contrast baseline, the predicted synthetic time-series, the corresponding pixel-wise uncertainty maps ($\sigma$), and the ground truth (GT) time-series. The kinetic plots illustrate the enhancement dynamics, comparing the predicted mean tumor intensity against the ground truth over time. The consensus tumor characterization across synthetic and real images, and corresponding diagnostic impact are summarized for each case in text format.}
    \label{fig:reader_study}
\end{figure*}

\subsection{Clinical Reader Study} \label{sec:reader_study}
To systematically evaluate the diagnostic equivalence,
realism, and clinical utility of the synthetic contrast-enhanced MRI images, we conducted a comprehensive multi-part clinical reader study with four radiologists, with 5, 8, 13, and 15 years of experience respectively, evaluating randomly selected forty cases across three stratification variables: tumor size, tumor shape and enhancement pattern. The study design, reader profile, statistical tests and additional results are added in Appendix \ref{appendix:reader_study}.  Readers can directly access platform here: \url{https://smriti-joshi.github.io/reader-study-contrast-synthesis/}.

\textit{Which method do the readers prefer?}

In a direct comparative evaluation against established temporal contrast synthesis baselines (UNet and TeNCA), readers demonstrated an strong preference for the proposed method. Specifically, out of 160 total evaluations (40 cases x 4 readers), our method was selected as the superior image in 83.8\% of cases (134 votes). In contrast, the compared architectures tied for a second, with UNet and TeNCA each capturing 8.1\% (13 votes) of the total preference. Statistical analysis confirmed that this dominance was highly significant ($p < 0.001$) and not due to chance.  This preference for our method was consistent across all four individual readers, while the two alternative methods were statistically indistinguishable from one another ($p = 1.0$).





\textit{How do real and synthetic image characteristics compare?}

The reader study comparing synthetic and ground truth images demonstrated a robust overall mean agreement rate of 69.8\% (median 66.7\%). As seen in Table \ref{tab:char}, coarse categorical judgments showed substantial GT-to-synthetic agreement, whereas margin and internal-enhancement assessments were only moderate. Readers noted a divergence in ordinal assessments such as image quality (56.2\% agreement) and kinetic plausibility (48.8\%), where real images maintained a clear scoring advantage. Nonetheless, for both criterion, synthetic and real images are scored between 2 and 3, suggesting that the readers found the kinetic curves plausible and the image quality acceptable on average (further demonstrated in Appendix Figure \ref{appendix_fig:composite_results}).

\begin{table}[t]
\scriptsize
\centering
\caption{\textbf{Lesion characterization: GT vs.\ synthetic.} Top: categorical
tasks (Cohen's $\kappa$ with bootstrap 95\% CI). Bottom: ordinal tasks
(GT/synthetic means, mean difference $\Delta$, signed rank-biserial $r$, Wilcoxon $p$). All $p$-values survive Benjamini--Hochberg correction.}
\label{tab:char}
\begin{tabular}{lcccc}
\toprule
\rowcolor{pink!25}
\textbf{Categorical task} & \textbf{N} & \textbf{\% agree} & \textbf{Cohen's $\kappa$} & \textbf{95\% CI} \\
\midrule
Lesion Type & 160 & 86.2 & 0.72 & [0.62, 0.83] \\
Shape & 119 & 89.9 & 0.73 & [0.57, 0.86] \\
Margins & 153 & 77.1 & 0.45 & [0.30, 0.59] \\
Enhancement & 160 & 68.8 & 0.43 & [0.31, 0.55] \\
\bottomrule
\end{tabular}

\vspace{0.6em}

\begin{tabular}{lccccc}
\toprule
\rowcolor{pink!25}
\textbf{Ordinal task} & \textbf{GT} & \textbf{Synth} & \textbf{$\Delta$} & \textbf{$r$} & \textbf{$p$} \\
\midrule
Kinetic Plausibility & 2.71 & 2.27 & -0.43 & -0.62 & $<$0.001 \\
Image Quality & 2.81 & 2.37 & -0.44 & -0.77 & $<$0.001 \\
\bottomrule
\end{tabular}
\end{table}

\textit{How does synthetic data impact patient care?}

\begin{table}[t]
\scriptsize
\centering
\caption{\textbf{Predictors of clinical impact.} Proportional-odds ordinal
logistic regression; standardized predictors. OR $>1$ indicates higher odds of
greater clinical impact.}
\label{tab:ordreg}
\begin{tabular}{lrrl}
\toprule
\rowcolor{pink!25}
\textbf{Predictor} & \textbf{OR} & \textbf{95\% CI} & \textbf{$p$} \\
\midrule
Perceived synthetic appearance & 3.60 & 2.26--5.74 & $<$0.001 \\
Diagnostic complexity & 2.14 & 1.46--3.12 & $<$0.001 \\
Image quality & 0.86 & 0.56--1.33 & 0.507 \\
Reader experience & 0.54 & 0.38--0.77 & $<$0.001 \\
Tumor size (large) & 0.82 & 0.59--1.14 & 0.235 \\
Shape (irregular) & 1.24 & 0.89--1.72 & 0.197 \\
Peak enhancement (late) & 0.84 & 0.60--1.18 & 0.318 \\
\bottomrule
\end{tabular}
\end{table}

In the side-by-side review of real and synthetic images, readers judged that relying on the synthetic image would produce a major change in clinical management in 30.0\% of assessments. However, 34.4\% of evaluations resulted in minor diagnostic deviations that did not alter the clinical management plan, and 35.6\% resulted in no change at all. In other words, in 70.0\% of the cases evaluated, replacing real images with synthetic ones would not negatively alter patient care. Further analysis using a proportional-odds ordinal logistic regression model identified perceived synthetic appearance (OR 3.60, $p < 0.001$) and diagnostic complexity (OR 2.14, $p < 0.001$) as significant independent predictors of greater clinical impact (i.e., a higher likelihood of altered patient management). Conversely, higher reader work experience served as a significant protective factor against these deviations (OR 0.54, 95\% CI 0.38--0.77, $p < 0.001$). Image quality was not independently predictive of clinical impact once perceived appearance and complexity were accounted for in the model (OR 0.86, $p = 0.51$; Table \ref{tab:ordreg}).

\textit{How does uncertainty map affect reader confidence?}

When presented with synthetic image and corresponding uncertainty maps, readers indicated that they would avoid or defer physical contrast injection in 64\% of the evaluated cases. To help guide these critical management decisions, the integration of uncertainty served as a valuable, though imperfect, adjunct. This map was computed by adding random noise for 10 different iterations to produce slightly different outputs. As seen in Figure \ref{fig:reader_study}, the areas where the predictions agreed were displayed with high model confidence (black) while disagreement indicated uncertainty (yellow). In the reader study, map provided actionable guidance in 49\% of evaluations overall, meaningfully shifting radiologist confidence by either validating trustworthy synthetic images (increasing confidence in 32\% of cases) or flagging unreliable ones (decreasing confidence in 17\%). These shifts directly influenced downstream actions: when the map decreased confidence, it acted as an effective safety mechanism, prompting readers to request true contrast injection in 88\% of those instances. Conversely, when it increased confidence in unchanged assessments, 82\% of readers proceeded without requesting real contrast.

Furthermore, statistical analysis revealed that the map's directional influence is significantly modulated by case complexity (Kendall's $\tau = 0.218$, $p = 0.003$). While the map predominantly increased confidence in low-complexity cases (39\% increase vs. 11\% decrease), it functionally inverted in high-complexity scenarios, where it more frequently decreased confidence (41\% decrease vs. 14\% increase) to flag unreliable generations. Consequently, its overall informative rate peaked during these highly complex evaluations (55\%). This utility also demonstrated a trending association with lesion type ($p = 0.059$), proving especially beneficial for challenging non-mass enhancement (NME) lesions, which achieved a 67\% informative rate. However, these metrics must be interpreted with distinct clinical caution. Inter-reader agreement regarding the map's effect was notably low (Fleiss' $\kappa = 0.02 \text{--} 0.10$), indicating that reliance on the map remains highly subjective and reader-dependent. Furthermore, qualitative feedback highlighted specific limitations with the map's reliability, noting critical instances where it incorrectly displayed high confidence despite the underlying synthetic image being clinically inaccurate. This susceptibility to overconfident errors underscores that while the uncertainty map is a helpful decision-support tool in the aggregate, it is not a definitive fail-safe, and must continue to be scrutinized carefully to prevent diagnostic missteps.

\section{Discussion and Conclusion}
\label{sec:conclusion}

This work presents a conditioned latent transport framework for single-step DCE-MRI contrast synthesis. Quantitative evaluations demonstrate that the approach effectively balances spatial fidelity with precise pharmacokinetic temporal alignment, outperforming baseline models on internal data. Independent external validation evaluates the model's adaptability to varying scanner noise profiles and temporal acquisition shifts. While the framework demonstrates promising resilience, the performance degradation under differing clinical protocols indicates that cross-institutional generalization remains an ongoing challenge

Furthermore, the ablation studies evaluate the design choices required for temporal contrast synthesis. Specifically, anchoring to a pre-contrast image enforces physiological adherence, allowing the model to focus on temporal synthesis, individual loss components improve visual and temporal fidelity, while a fixed noise strategy maintains continuity in densely sampled predictions.

In the downstream tumor segmentation task, our proposed framework outperforms competing state-of-the-art methods across two distinct evaluation strategies. Success on a network natively trained on post-contrast images confirms that the synthesized contrast uptake accurately mirrors the ground truth. Superior performance on a pre-contrast trained network demonstrates that the synthesized data strictly preserves the underlying morphological and textural characteristics. These quantitative successes translate directly to clinical viability. In a blinded reader study conducted by expert breast radiologists, our synthesized images were preferred over competing baselines in approximately 84\% of cases. Highlighting the framework's potential to safely reduce contrast burden, the study further confirmed that substituting real post-contrast MRI with our synthetic counterparts would result in no major negative deviation to patient management in 70\% of cases.

Qualitative feedback from the reader study indicates clinical optimism regarding the potential of synthetic contrast-enhanced MRI to reduce patient contrast burden. At the current maturity level of the state-of-the-art in contrast synthesis, conventional DCE-MRI remains indispensable for precise preoperative surgical planning. However, radiologists acknowledged that upon rigorous clinical validation, the technology shows promise as a diagnostic adjunct. Specifically, in their opinion, integrating synthetic contrast with high-resolution diffusion-weighted imaging (DWI) or mammography could elevate baseline screening confidence without necessitating immediate contrast administration. This aspect is not explicitely tested in this study and can make for an interesting future direction. In addition, further model refinements must address specific morphological and kinetic blind spots. Targeted improvements should prioritize enhancing margin sharpness in dense breast tissue, accurately reproducing internal tumor heterogeneity (e.g., central necrosis), and reliably capturing the true spatial extent of small lesions and non-mass enhancement (NME). Furthermore, it is important to increase robustness in failure detection through uncertainty estimation and other complementary methods. We add one failure case from the reader study in Appendix Figure \ref{fig:failure_clinical_study} which obtained identical diagnostic assessment, but the tumor is completely mislocalized.

Strictly in terms of methodology, our work has several limitations. First, our current pipeline relies on qualitative assessments of pre- and post-contrast image registration, meaning that patient motion between acquisitions was not strictly quantified or corrected. Second, the training dataset exhibits a long-tailed temporal distribution, with sparse representation of early ($<100\text{ s}$) and late ($>500\text{ s}$) acquisition phases (Appendix Figure \ref{appendix_fig:data_distribution}). Consequently, the dataset contains fewer examples of tumors captured during the rapid wash-in period or exhibiting late-phase enhancement. We explored sampling techniques to address this imbalance, but they failed to improve synthesis quality in these low-density temporal regions without simultaneously degrading performance on the broader test set (Appendix Table \ref{appendix_tab:sampling}). Addressing this domain gap to ensure kinetic fidelity in low-data regimes remains a critical area for future investigation. Third, in our current formulation, the model's integration timestep $t$ and the physical acquisition time $\tau$ are fully decoupled. Given the highly non-linear nature of contrast uptake in malignant tumors, characterized by a rapid wash-in phase and a gradual wash-out phase, future work could explicitly couple these variables by utilizing a physiological pharmacokinetic (PK) model as the interpolation function. While a strict PK-driven trajectory could enforce rigorous physical priors in highly controlled, single-center protocols, it presents a unique challenge for generalized models. Because the MAMA-MIA dataset aggregates data from over 25 institutions with widely heterogeneous injection protocols and temporal resolutions, enforcing a single, idealized PK curve could act as an overly restrictive inductive bias. Future architectures must balance these physiological priors with the flexibility required to model multi-center clinical variance. Finally, owing to limited data availability, we did not explicitly evaluate synthetic contrast uptake in benign lesions. Because benign and malignant masses typically exhibit distinct pharmacokinetic enhancement profiles, extending this framework to reliably model and differentiate between these varying pathologies remains an essential objective for future clinical translation.

In conclusion, the aim of this work is to contextualize the current progress of contrast synthesis breast DCE-MRI. We hope that our evaluation framework serves an anchor for future research, ensuring that progress in this domain remains transparent and clinically meaningful. We also hope that the community builds upon the gaps identified in the study to advance temporal contrast generation.

\printcredits

\section*{Acknowledgements}
This project has received funding from the European Union's Horizon 2020 research and innovation programme under grant agreement No 101057699 (RadioVal). Additionally, this work was partially supported by the project FUTURE-ES (PID2021-126724OB-I00) and project AIMED (PID2023-146786OB-I00) from the Ministry of Science and Innovation of Spain.
D.M.L. and J.A.S. received funding from HELMHOLTZ IMAGING, a platform of the Helmholtz Information and Data Science Incubator. G.S. received funding from the European Union's Horizon Europe research and innovation programme under Grant Agreement No. 101057849 (DataTools4Heart).

The authors acknowledge the role of Gemini 3.1 Pro for enhancing clarity in the text, as well as Claude Sonnet 4.6 and DeepSeek V4 Flash for assistance in coding. After using these tools, the authors reviewed and edited the content as needed and take full responsibility for the content of the published article.

\section{Ethical Approval Statement}
For the publicly available datasets used in this study, formal ethics committee approval and written informed consent were not required. For external validation data, ethical approval was obtained from the Swedish Ethical Review Authority (Etikprövningsmyndigheten) (Approval Number: 2020-00488, with subsequent amendment 2022-06777-02). The requirement for written informed consent was waived due to the retrospective nature of the study.
\section{Declaration of competing interest}
Given their role as Medical Image Analysis Associate Editor, Julia Schnabel had no involvement in the peer-review of this article and has no access to information regarding its peer-review. Full responsibility for the editorial process for this article was delegated to another journal editor. The authors declare that they have no additional competing financial
interests or personal relationships that could have appeared to influence
the work reported in this paper.

\section{Data Availability}
The MAMA-MIA dataset~\citep{garrucho2025large} and DUKE-Breast-Cancer-MRI dataset~\citep{duke} are publicly available on Synapse\footnote{https://www.synapse.org/Synapse:syn60868042/wiki/628716} and TCIA\footnote{https://www.cancerimagingarchive.net/collection/duke-breast-cancer-mri/}, respectively. The external validation KI dataset is private and the authors do not have permission to release it.

\section{Code Availability}
The source code will be made publicly available upon acceptance.

\begin{appendices}

\renewcommand{\thetable}{\thesection.\arabic{table}}
\renewcommand{\thefigure}{\thesection.\arabic{figure}}
\setcounter{table}{0}
\setcounter{figure}{0}

\section{Proposed Model} \label{appendix:contrast_synthesis}

\begin{figure*}[p]
    \centering
    \includegraphics[width=0.9\linewidth]{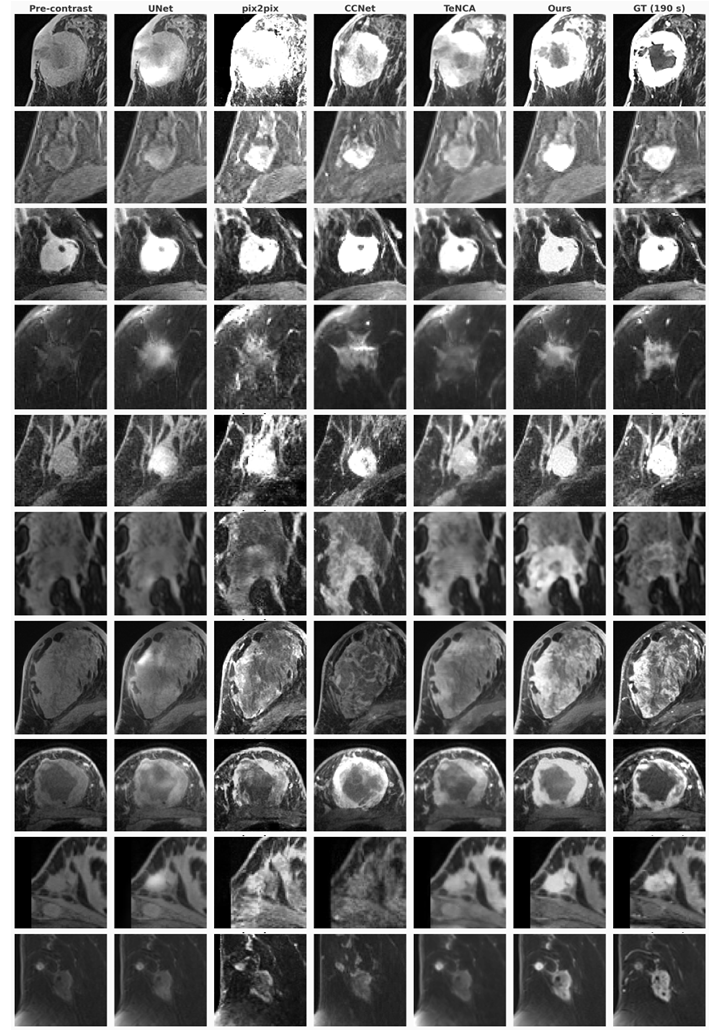}
    \caption{Additional Qualitative Results for Synthetic Contrast Generation.}
    \label{appendix_fig:qualitative}
\end{figure*}

The implementation details and hyperparameter configuration of the proposed architecture is detailed in this section. Figure~\ref{appendix_fig:qualitative} shows additional qualitative comparison between different methods. 

\subsection{Preprocessing}
This work follows the same preprocessing pipeline established in TeNCA~\citep{tenca}, with resampling to a uniform voxel spacing of 1 mm and intensity values are linearly rescaled between zero and one based on the 0.02 and 99.98 percentiles of the respective pre-contrast image. Additionally, images are cropped to patches of size 168$\times$168. 

\subsection{Network and Hyperparameters} 
\textit{Variational autoencoder:} The model is trained with a composite loss function with MSE weighted at 1.0, perceptual (LPIPS) loss weighted at 5.0, complemented by a small KL regularization term (1e-6), a Sobel gradient-based edge-sharpness loss (0.5). Training runs for 200 epochs with a batch size of 16, a learning rate of 1e-4, and gradient clipping at 1.0.  This yields a 4-channel latent space derived from a VAE with a $4\times$ spatial compression and a latent scaling factor of 1.0259.

\textit{Latent UNet:} The core latent space architecture utilizes DiffusionModeUNet (from MONAI Generative \citep{pinaya2023generative}), with channel configuration of $[128, 256, 512, 512]$, incorporating two residual blocks per resolution stage and spatial self-attention at the three highest-resolution levels. The noise level fixed at $\sigma = 0.1$. We employ classifier-free guidance, applying a conditioning dropout rate of 0.1 during training and a guidance scale of 1.0 during inference. Training is conducted over 200 epochs with a batch size of 8 using the AdamW optimizer (learning rate: $1 \times 10^{-4}$, weight decay: $3 \times 10^{-4}$). To ensure training stability and efficiency, we apply gradient clipping at 1.0 and utilize mixed-precision training. The composite objective function comprises an MSE loss in the latent space ($\lambda = 1.0$), a perceptual loss ($\lambda = 5.0$), and a focal frequency loss ($\lambda = 50$). Finally, an exponential moving average (EMA) of the network weights is maintained with a decay rate of 0.999.

\subsection{The Impact of VAE reconstruction} \label{appendix:vae}                                                                                 
Table~\ref{appendix_tab:vae_ablation} evaluates the effect of the autoencoder (VAE) architecture on reconstruction fidelity (Phase 1) and temporal contrast synthesis (Phase 2). Using a pretrained autoencoder (Stability VAE, 8x downsampling)\footnote{https://huggingface.co/stabilityai/sd-vae-ft-mse} creates a spatial information bottleneck. Despite medical-domain fine-tuning, the 8x compression degrades high-frequency anatomical structures, resulting in a Phase 1 SSIM of 0.79. A custom VAE, trained from scratch with a 4x downsampling factor, preserves these details and improves structural reconstruction (SSIM: 0.92, PSNR: 30.80).

This reconstruction performance directly impacts the Phase 2 generative temporal synthesis. The 4x latent space retains finer details, enabling the network to map the pharmacokinetic trajectory while maintaining spatial boundaries. This yields improved pixel fidelity (MSE: $0.84 \times 10^{-2}$) and spatiotemporal alignment across the full image and isolated lesions (DTW: 0.68, DTW-ROI: 3.76). The 8x latent space yields better perceptual deep-feature scores (FID-Dinov2: 131.19 vs. 154.56), likely due to its closer alignment with the natural-image distributions expected by the pre-trained metric. However, the 4x architecture performs better on temporal and localized metrics, including FRD (4.98 vs. 5.72).

\begin{table}
\scriptsize
\centering
\caption{\textbf{Ablation Study.} Comparing autoencoder reconstruction fidelity (Phase 1) and its downstream impact on temporal synthesis (Phase 2). Best results are highlighted in \textbf{bold}.  MSE is reported in $10^{-2}$ scale.}
\label{appendix_tab:vae_ablation}
\begin{tabular}{lccc}
\toprule
\rowcolor{pink!25}
\textbf{Metric} & \textbf{\shortstack{Stability VAE \\ (Zero-Shot)}} & \textbf{\shortstack{Stability VAE \\ (Fine-tuned)}} & \textbf{\shortstack{VAE 4x \\ (Scratch)}} \\ \midrule
\multicolumn{4}{c}{\textbf{Phase 1: Pure VAE Reconstruction}} \\ \midrule
MSE $\downarrow$     & 0.33 (0.21) & 0.31 (0.20) & \textbf{0.11 (0.08)} \\
PSNR $\uparrow$      & 25.82 (3.25) & 26.13 (3.27) & \textbf{30.80 (3.14)} \\
SSIM $\uparrow$      & 0.78 (0.07) & 0.79 (0.07) & \textbf{0.92 (0.04)} \\
LPIPS $\downarrow$   & 0.12 (0.02) & 0.11 (0.02) & \textbf{0.10 (0.03)} \\
FID-Dinov2 $\downarrow$     & 79.13 & \textbf{78.18} & 120.99 \\ \midrule
\multicolumn{4}{c}{\textbf{Phase 2: Synthesis}} \\ \midrule
MSE $\downarrow$      & -- & 0.97 (0.60) & \textbf{0.84 (0.55)} \\ 
PSNR $\uparrow$& -- & 20.93 (2.67) & \textbf{21.61 (2.77)} \\
SSIM $\uparrow$     & -- & 0.70 (0.11)& \textbf{0.71 (0.11)} \\
LPIPS $\downarrow$      & -- & \textbf{0.18 (0.05)} & \textbf{0.18 (0.05)} \\
FID-Dinov2 $\downarrow$      & -- & \textbf{131.19} & 154.56  \\
\textbf{FRD $\downarrow$}  & -- & 5.72 & \textbf{4.98} \\
\textbf{DTW $\downarrow$} & -- & 0.74 (0.65) & \textbf{0.68 (0.56)} \\
\textbf{DTW-ROI $\downarrow$} & -- & 4.07(3.72) & \textbf{3.76 (3.33)} \\
\bottomrule
\end{tabular}%
\end{table}

\section{Segmentation} \label{appendix:segmentation}
\begin{figure*}[pH]
    \centering
    \includegraphics[width=0.9\linewidth]{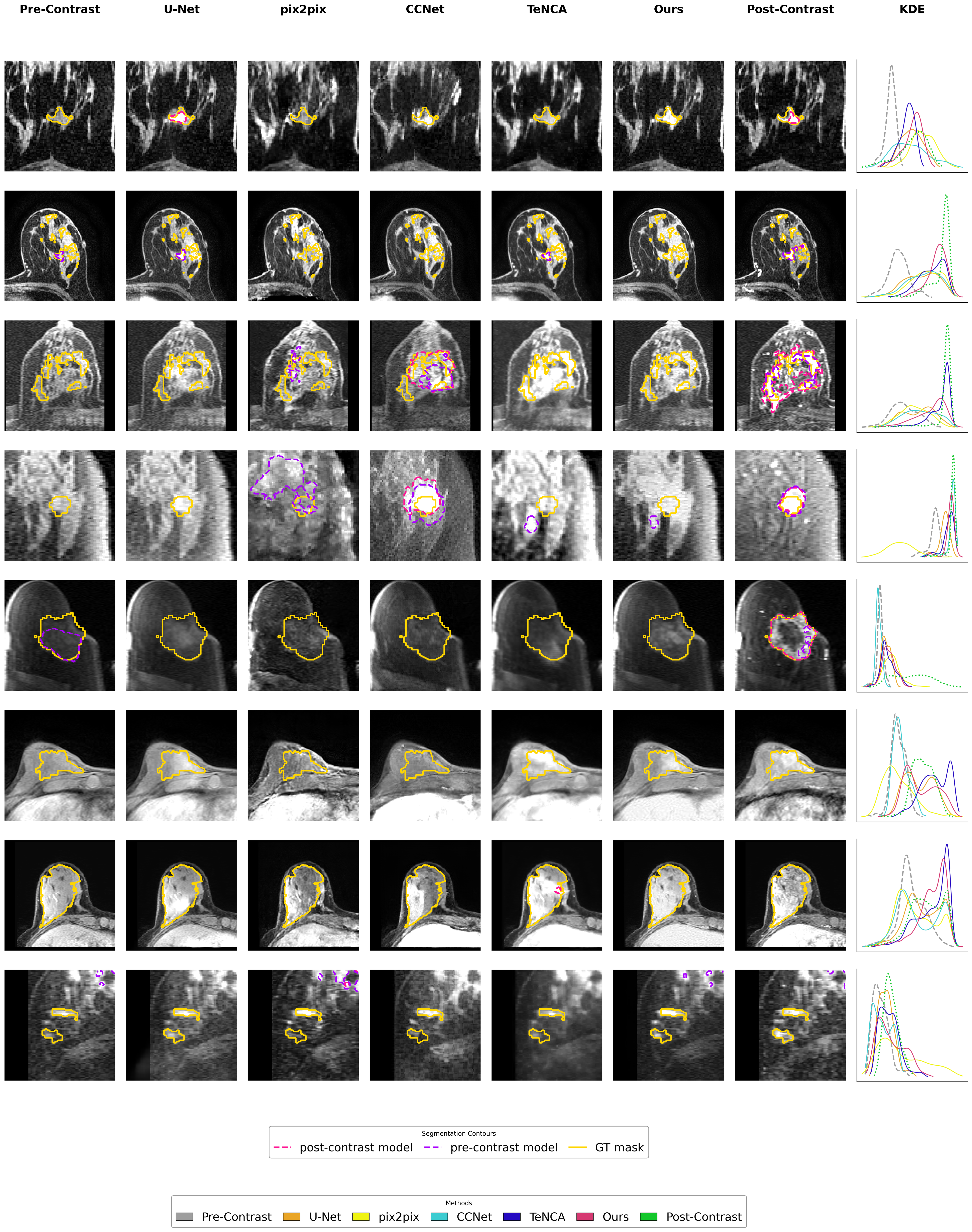}
    \caption{Failure Cases in Downstream Tumor Segmentation Task.}
    \label{appendix_fig:segmentation_failure}
\end{figure*}

\subsection{Training Details}
For downstream segmentation, we trained a 2D network using the standard nnUNet framework \citep{isensee2021nnu}, utilizing both the pre-contrast and first post-contrast images of the MAMA-MIA training set. The automated configuration heuristic generated a deep 2D U-Net architecture tailored to the median image dimensions and spacing of the dataset. Preprocessing included resampling all modalities to an in-plane resolution of 0.703 $\times$ 0.703 mm (using third-order spline interpolation for image data and nearest-neighbor for segmentation masks) alongside global Z-score normalization. Model optimization was executed on 256 $\times$ 256 pixel patches using a batch size of 49 and a batch-aggregated Dice loss function.

\subsection{Performance on all post contrast phases}
Table~\ref{tab:segmentation_internal_all_sequences} extends the downstream segmentation evaluation presented in the main manuscript by reporting performance across all synthesized post-contrast sequences, rather than exclusively the first acquisition phase. This extended analysis provides strong evidence that our generative framework successfully synthesizes an identifiable and persistent contrast enhancement trajectory as temporal acquisition progresses.

However, we note an expected slight degradation in the absolute metrics, including the real post-contrast Upper Bound, which decreases from a Dice of 0.68 to 0.62 on the post-contrast trained network. This is due to a structural limitation in the evaluation paradigm: the reference ground truth (GT) segmentation masks were clinically delineated based strictly on the first post-contrast sequence. As the contrast agent naturally diffuses into surrounding tissues during later acquisition phases, the physiological enhancement boundaries shift relative to this static, early-phase GT mask. Consequently, the multi-phase metrics reported here accurately reflect sustained temporal contrast identifiability, but serve only as a proxy for absolute boundary precision in late-phase dynamics.

\subsection{External Validation} 
Table~\ref{appendix_tab:seg_externalval} presents the downstream tumor segmentation performance evaluated on the independent, multi-institutional external validation cohort. As anticipated, the absolute quantitative metrics are lower compared to internal validation, including the real post-contrast upper bound (Dice: 0.60), which reflects the inherent domain gap when applying a fixed segmentation network to external data. Despite this overall reduction in absolute performance, the relative hierarchical trends established in the internal validation remain consistent, where our proposed method maintained robust biological feature amplification, significantly outperforming all synthetic baselines (Dice: 0.43 and 0.46 with post-contrast and pre-contrast trained segmentation networks respectively).

\begin{table*}
\scriptsize
\centering
\caption{Downstream Segmentation Performance on Internal Validation Data (extended with evaluation on all sequences). Baseline and Upper Bound refer to inference on pre-contrast images and real post-contrast images, respectively.  Statistical significance was computed via a paired Wilcoxon signed-rank test ($p < 0.01$). Results that did not reach statistical significance are marked with an asterisk ($*$).}
\label{tab:segmentation_internal_all_sequences}
\begin{tabular}{lcccc}
\toprule
\rowcolor{pink!25}
\textbf{Method} & \multicolumn{2}{c}{First} & \multicolumn{2}{c}{All} \\
\rowcolor{pink!25}
 & \multicolumn{2}{c}{Post-Contrast} & \multicolumn{2}{c}{Post-Contrast} \\
\cmidrule(lr){2-3}\cmidrule(lr){4-5}
 & \textbf{Dice} $\uparrow$ & \textbf{HD95} $\downarrow$ & \textbf{Dice} $\uparrow$ & \textbf{HD95} $\downarrow$ \\
\midrule

\multicolumn{5}{@{}l}{\textit{Segmentation model trained on post-contrast}} \\ \midrule
Baseline     & 0.17 (0.30) & 160.69 (104.48) & 0.17 (0.30) & 160.69 (104.48) \\
Upper Bound & 0.68 (0.33) & 38.59 (77.73)  & 0.62 (0.32) & 51.91 (79.03) \\
U-Net                       & 0.44 (0.36) & 82.20 (104.68)* & 0.44 (0.35) & 83.32 (98.23)* \\
pix2pix  & 0.22 (0.33) & 147.79 (108.24) & 0.22 (0.33) & 147.79 (108.24) \\
CCNet        &  0.45 (0.35)           &  76.81 (102.32)  & 0.40 (0.30) &   84.67 (87.48) \\
TeNCA        & 0.30 (0.36) & 122.31 (110.69) & 0.25 (0.32) & 145.04 (94.63) \\
\textbf{Ours}               & \textbf{0.52 (0.36)} & \textbf{63.07 (95.85)} & \textbf{0.48 (0.36)} & \textbf{75.48 (97.23)} \\
\midrule

\multicolumn{5}{@{}l}{\textit{Segmentation model trained on pre-contrast}} \\ \midrule
Baseline      & 0.49 (0.37) & 71.48 (99.90) & 0.49 (0.37) & 71.48 (99.90) \\
Upper Bound  & 0.63 (0.34) & 47.46 (84.74) & 0.63 (0.31) & 41.33 (71.74) \\
U-Net                       & 0.56 (0.35) & 53.05 (89.01) & 0.57 (0.33) & 49.98 (83.36) \\
pix2pix & 0.44 (0.35) & 70.10 (96.55) & 0.44 (0.35) & 70.09 (96.55) \\
CCNet        & 0.44 (0.35) & 78.51 (102.96) & 0.46 (0.31) &  67.66 (83.13) \\
TeNCA          & 0.51 (0.36) & 62.59 (93.86) & 0.53 (0.34) & 59.52 (86.86) \\
\textbf{Ours}               & \textbf{0.60 (0.33)} & \textbf{43.38 (80.12)} & \textbf{0.60 (0.33)} & \textbf{44.26 (79.37)} \\
\bottomrule
\end{tabular}
\end{table*}

\begin{table*}
\scriptsize
\centering
\caption{Downstream Segmentation Performance on External Validation Data. Baseline and Upper Bound refer to inference on pre-contrast images and real post-contrast images, respectively.  Statistical significance was computed via a paired Wilcoxon signed-rank test ($p < 0.01$). Results that did not reach statistical significance are marked with an asterisk ($*$).}
\label{appendix_tab:seg_externalval}
\begin{tabular}{lcccc}
\toprule
\rowcolor{pink!25}
\textbf{Method} & \multicolumn{2}{c}{First} & \multicolumn{2}{c}{All} \\
\rowcolor{pink!25}
 & \multicolumn{2}{c}{Post-Contrast} & \multicolumn{2}{c}{Post-Contrast} \\
\cmidrule(lr){2-3}\cmidrule(lr){4-5}
 & \textbf{Dice} $\uparrow$ & \textbf{HD95} $\downarrow$ & \textbf{Dice} $\uparrow$ & \textbf{HD95} $\downarrow$ \\
\midrule
\multicolumn{5}{@{}l}{\textit{\textit{Segmentation model trained on post-contrast}}} \\ \midrule
Baseline     & 0.07 (0.21) & 201.92 (81.43) & 0.07 (0.21) & 201.92 (81.43) \\
Upper Bound & 0.60 (0.35) & 53.55 (90.56)  & 0.54 (0.33) & 67.61 (89.76) \\
U-Net                       & 0.38 (0.33)* & 84.19 (105.11)* & 0.38 (0.31)* & 83.30 (97.52)* \\
pix2pix & 0.09 (0.23) & 191.52 (89.97) & 0.09 (0.23) & 191.52  (89.96) \\
CCNet  &0.36 (0.32) &  76.45 (101.29) &  0.33 (0.28) &  89.29 (81.12)  \\
TeNCA          & 0.19 (0.30) & 152.64 (107.67) & 0.19 (0.29) & 153.31 (97.06) \\
\textbf{Ours}               & \textbf{0.43 (0.33)} & \textbf{59.28 (90.86)} & \textbf{0.40 (0.32)} & \textbf{76.32 (91.76)} \\
\midrule
\multicolumn{5}{@{}l}{\textit{Segmentation model trained on pre-contrast}} \\ \midrule
Baseline    & 0.36 (0.35) & 107.26 (110.60) & 0.36 (0.35) & 107.26 (110.60) \\
Upper Bound  & 0.51 (0.37) & 76.18 (101.95) & 0.52 (0.33) & 65.45 (86.69) \\
U-Net                       & 0.39 (0.34) & 88.55 (105.97) & 0.41 (0.33) & 77.12 (94.96) \\
pix2pix & 0.33 (0.31) & 95.46 (107.05) & 0.33 (0.31) & 95.46 (107.05) \\
CCNet         & 0.34 (0.33) & 93.74 (107.18) & 0.37 (0.31) &  82.77 (86.70)   \\
TeNCA        & 0.41 (0.33) & 80.45 (101.59) & 0.37 (0.30) & 93.90 (87.15) \\
\textbf{Ours}               & \textbf{0.46 (0.33)} & \textbf{65.86 (95.69)} & \textbf{0.46 (0.32)} & \textbf{62.07 (90.57)} \\
\bottomrule
\end{tabular}
\end{table*}

\subsection{Failure Cases} \label{appendix:segmentation_failure}
An analysis of complete segmentation failures (Dice = 0) further highlights the robustness of our method. Natively trained post-contrast networks proved sensitive to any deviation from post-contrast distribution, yielding 238 unique patient failures across all methods, compared to  150 for the robust pre-contrast network. As seen in Table~\ref{appendix_tab:zero_dice_failures}, our proposed framework exhibited the lowest catastrophic failure rate (46 cases), and even surpassing the real post-contrast ground truth (54 cases) when tested with pre-contrast segmentation network. Overall, we have the lowest failure rate across synthesis methods where 29/300 case were missed by both segmentation networks. We add representative failure cases of the proposed method in Figure~\ref{appendix_fig:segmentation_failure}.
\begin{table}
\scriptsize
\centering
\caption{Analysis of catastrophic segmentation failures (Dice = 0). The table reports the absolute number of unique patient cases where the segmentation networks completely failed to localize the lesion, evaluated across both pre-contrast and post-contrast training paradigms. "Overlap" indicates the number of cases that failed simultaneously under both paradigms.}
\label{appendix_tab:zero_dice_failures}
\begin{tabular}{lccc}
\toprule
\rowcolor{pink!25}
\textbf{Method} & \textbf{Pre-Contrast} & \textbf{Post-Contrast} & \textbf{Overlap} \\
\rowcolor{pink!25}
 & \textbf{Network Failures} & \textbf{Network Failures} & \textbf{(Both)} \\
\midrule
Pre-contrast     & 86 & 203 & 79 \\
Post-Contrast & \underline{54} & \textbf{40}  & \textbf{16} \\
U-Net                       & 60 & 96  & 45 \\
pix2pix  & 86 & 184 & 78 \\
CCNet  & 92 & 87 & 51 \\
TeNCA        & 63 & 153 & 53 \\
\textbf{Ours}               & \textbf{46} & \underline{82} & \underline{29} \\
\bottomrule
\end{tabular}
\end{table}

\begin{figure*}
    \centering
    \begin{subfigure}[b]{0.48\linewidth}
        \centering
        \includegraphics[width=\linewidth]{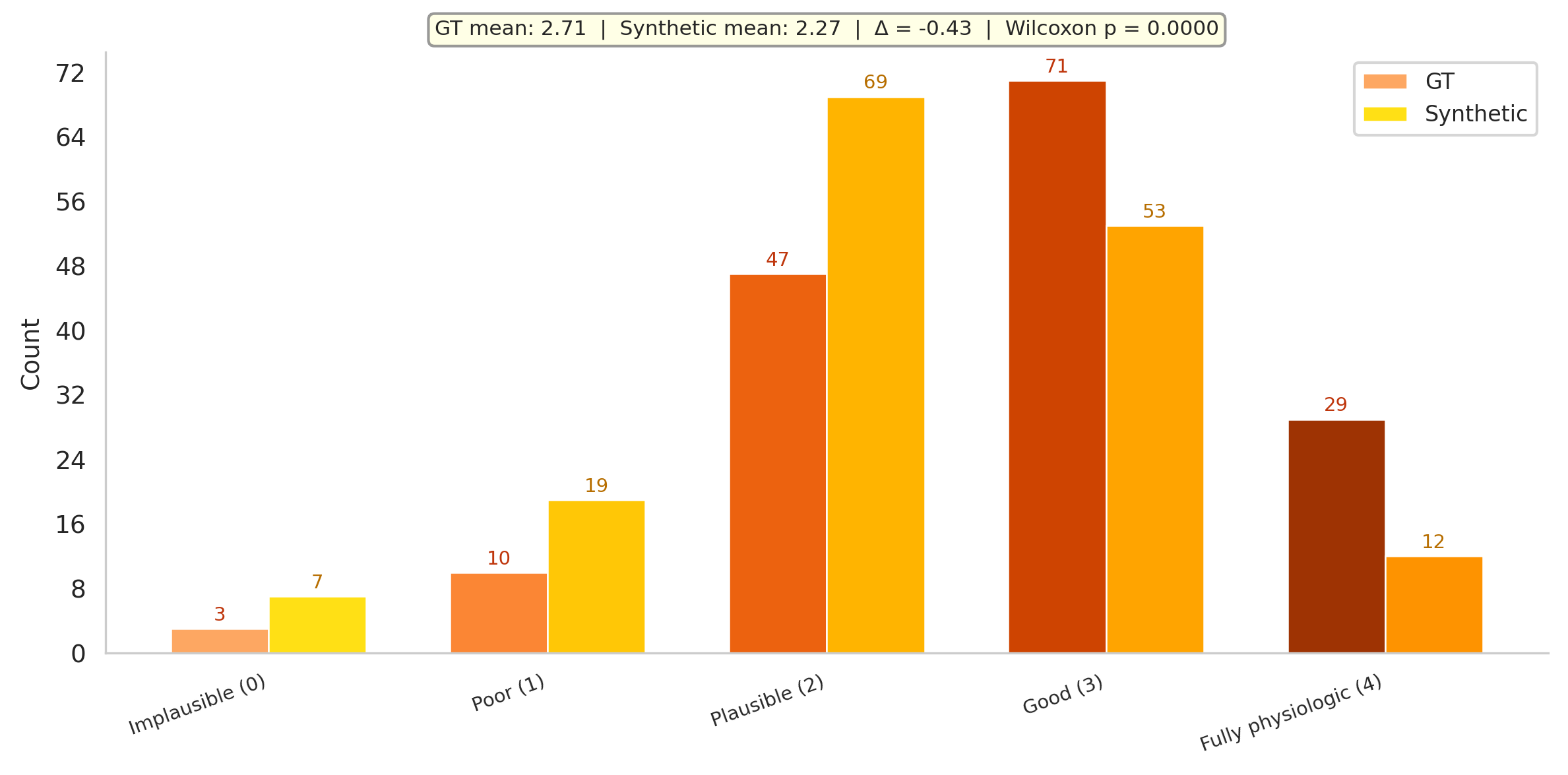}
        \caption{Paired Kinetic}
        \label{fig:paired_kinetic}
    \end{subfigure}
    \hfill
    \begin{subfigure}[b]{0.48\linewidth}
        \centering
        \includegraphics[width=\linewidth]{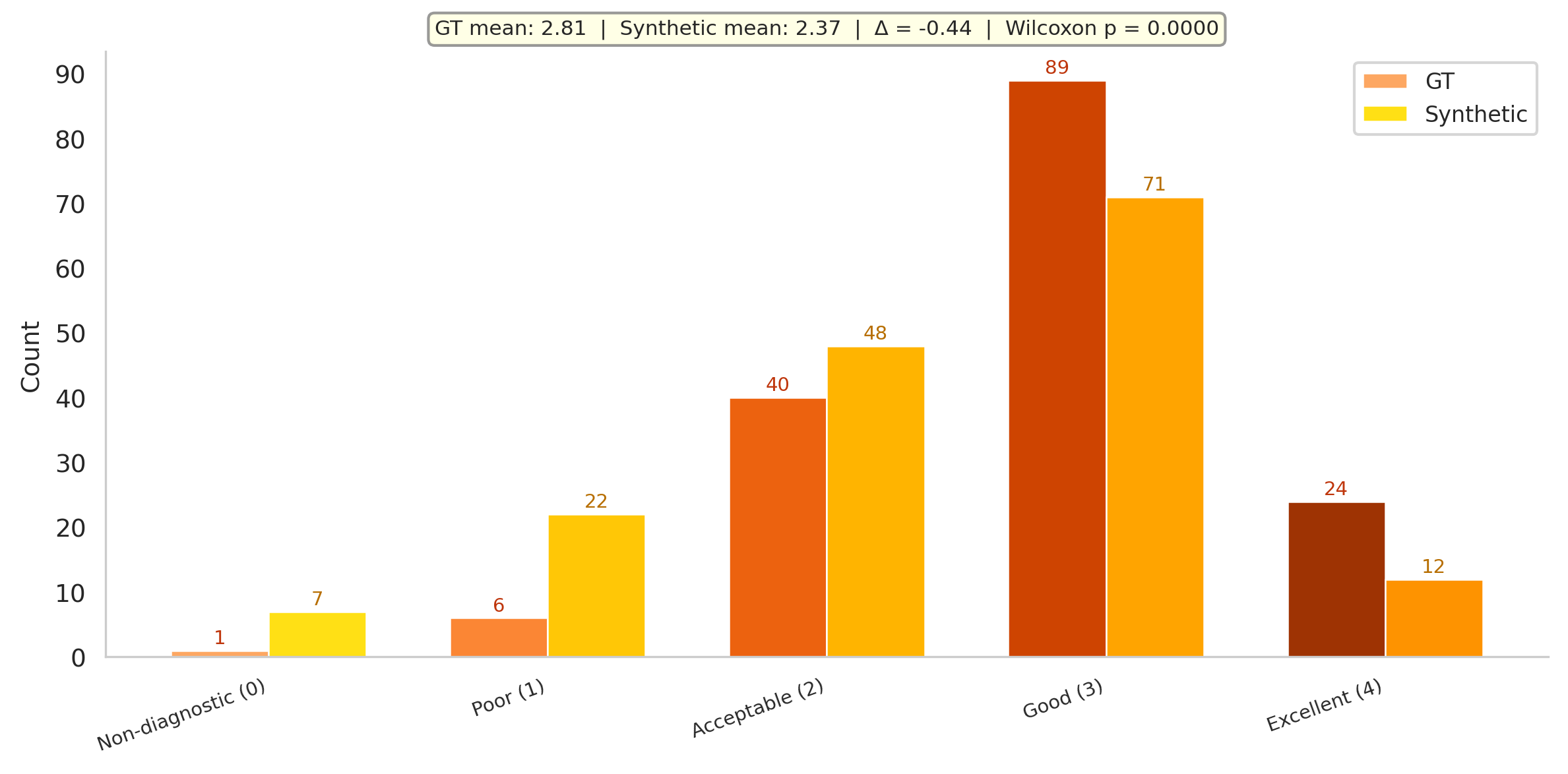}
        \caption{Paired Quality}
        \label{fig:paired_quality}
    \end{subfigure}
    
    \hfill
    
    \caption{Overall composite caption describing the kinetic analysis, quality metrics, and case agreement.}
    \label{appendix_fig:composite_results}
\end{figure*}

\begin{figure}
    \centering
    \includegraphics[width=\linewidth]{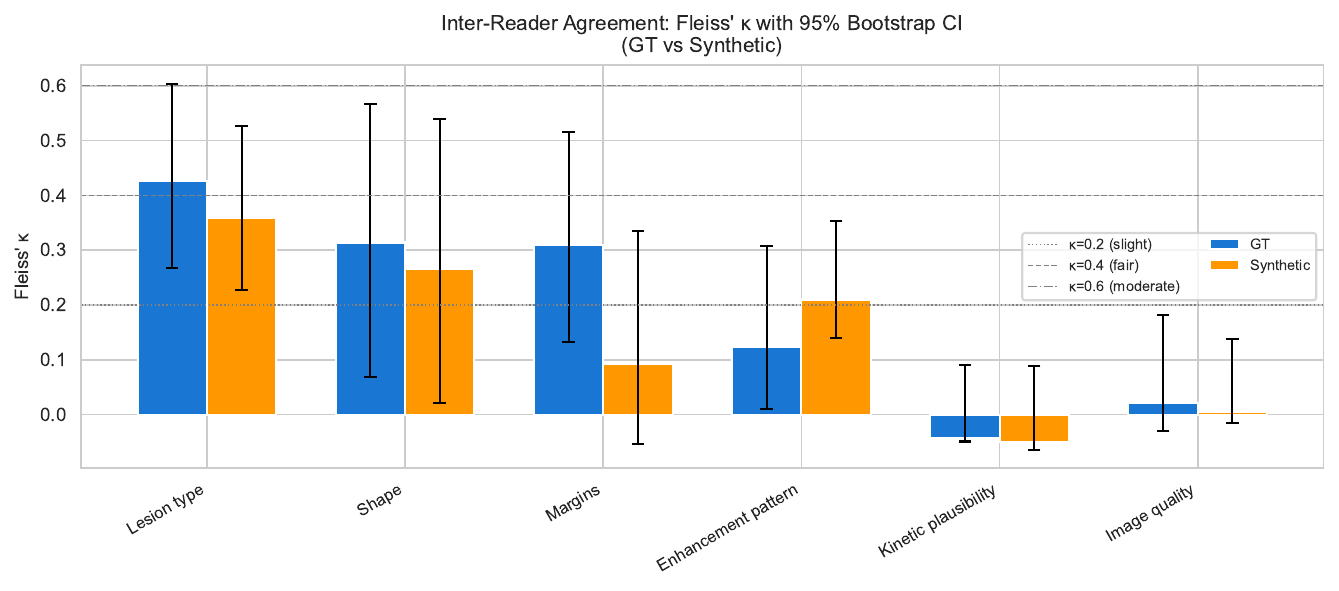}
    \caption{\textbf{Inter-reader agreement} on GT and synthetic images
    (Fleiss' $\kappa$ with bootstrap 95\% CI) across tasks.}
    \label{fig:interreader}
\end{figure}

\section{Clinical Reader Study} \label{appendix:reader_study}

To systematically evaluate the diagnostic equivalence, realism, and clinical utility of the synthetic contrast-enhanced MRI images, we conducted a comprehensive multi-part reader study. Readers can directly access platform here: \url{https://smriti-joshi.github.io/reader-study-contrast-synthesis/}.

\subsection{Study Design}

\subsubsection{Selection of Cases}
To ensure a balanced evaluation, a subset of 40 cases was selected from the internal test set using a stratified random sampling approach. Stratification was based on three objective metrics extracted directly from the ground-truth segmentations: tumor volume, boundary sphericity, and peak enhancement latency. To establish the phenotypic subgroups, each metric was dichotomized using a robust median split approach, where the median value across the entire test-set cohort served as the definitive cutoff threshold. Specifically, tumor volume was assessed using the maximal cross-sectional area in pixels, categorizing cases as Small (less than or equal to the median) or Large (greater than the median). Boundary sphericity was quantified using a standard circularity metric, mathematically defined as $\frac{4\pi \times \text{Area}}{\text{Perimeter}^2}$, with cases classified as Irregular (less than or equal to the median) or Regular (greater than the median). Finally, peak enhancement latency was measured as the acquisition time in seconds required for the mean signal intensity inside the tumor mask to reach its maximum across all post-contrast phases, categorizing cases as Early (less than or equal to the median) or Late (greater than the median). This categorization yielded eight distinct phenotypic subgroups. To prevent evaluation bias toward any single, frequently occurring tumor presentation, exactly five cases were randomly sampled from each of the eight subgroups. The detailed distribution is provided in Table~\ref{tab:case_selection}.

\begin{table*}
\scriptsize
\centering
\caption{Stratified case selection from the internal test set based on tumor volume, boundary sphericity, and peak enhancement latency.}
\label{tab:case_selection}
\begin{tabular}{lllcc}
\toprule
\rowcolor{pink!25}
\textbf{Tumor Volume} & \textbf{Boundary Sphericity} & \textbf{Peak Enhancement} & \textbf{Selected Cases} & \textbf{Total Pool} \\ 
\midrule
Large & Irregular & Early & 5 & 33 \\ 
Large & Irregular & Late  & 5 & 58 \\ 
Large & Regular   & Early & 5 & 34 \\ 
Large & Regular   & Late  & 5 & 24 \\ 
Small & Irregular & Early & 5 & 31 \\ 
Small & Irregular & Late  & 5 & 28 \\ 
Small & Regular   & Early & 5 & 52 \\ 
Small & Regular   & Late  & 5 & 39 \\ 
\bottomrule
\end{tabular}
\end{table*}

\begin{table}
\centering
\caption{\textbf{Paired vs.\ inter-reader agreement.} GT-to-synthetic Cohen's
$\kappa$ exceeds GT inter-reader Fleiss' $\kappa$ for every categorical
task.}
\label{tab:agree}
\begin{tabular}{lrrr}
\toprule
\rowcolor{pink!25}
\textbf{Task} & \textbf{GT$\leftrightarrow$Synth $\kappa$} & \textbf{Inter-reader $\kappa$} & \textbf{$\Delta$} \\
\midrule
Lesion Type & 0.72 & 0.43 & +0.30 \\
Shape & 0.73 & 0.31 & +0.41 \\
Margins & 0.45 & 0.31 & +0.14 \\
Enhancement & 0.43 & 0.12 & +0.31 \\
\bottomrule
\end{tabular}
\end{table}
\subsubsection{Lesion Characterization and Image Quality}
In the first section, readers were presented with isolated images (either real or synthetic, presented in a blinded fashion) and asked to characterize the breast tumor and assess the overall image quality. The characterization was based on the standard lexicon, requiring readers to classify the lesion type as a mass, non-mass enhancement (NME), focus/foci, or indeterminate. For lesions identified as masses, readers further specified the shape (round, oval, or irregular) and margins (circumscribed, irregular, spiculated, or indeterminate). The enhancement pattern was categorized as homogeneous, heterogeneous, rim enhancement, dark internal septations, or not assessable.

Beyond standard characterization, readers evaluated the kinetic plausibility of the enhancement using a 5-point scale ranging from 1 (implausible/non-physiologic pattern) to 5 (fully physiologic). Finally, readers rated the overall quality of the MRI image, independent of the tumor, on a 5-point Likert scale ranging from 1 (non-diagnostic due to severe artifacts) to 5 (excellent high-quality image with minimal artifacts).

\subsubsection{Clinical Decision-Making and Diagnostic Equivalence}
In the second section, readers performed a side-by-side comparative evaluation of the synthetic image against the corresponding real acquisition to determine diagnostic equivalence. Readers assessed the potential impact of the synthetic image on clinical decision-making, categorizing it as causing no change, a minor change (no impact on clinical management), or a major change (affecting clinical management or diagnosis). If a major change was indicated, readers specified the primary discrepancy, such as incorrect tumor localization, inaccurate extent/size estimation, implausible enhancement kinetics, image quality issues, or missed/false detections.

Additionally, readers rated the inherent diagnostic complexity of the case (low, moderate, or high) and evaluated the perceived synthetic appearance of the generated image on a 4-point scale from "None" (fully natural) to "Strong" (clearly synthetic). During this phase, readers were also provided with an AI-generated anomaly map indicating epistemic uncertainty. Readers reported how this maps influenced their diagnostic confidence (increased, decreased, or no change) and selected their next logical clinical step if presented with this data in a real workflow (e.g., proceed normally, review with caution, or request standard contrast injection).

\subsubsection{Method Preference and Comparative Benchmarking}
In the final phase, the proposed generation method was benchmarked against alternative state-of-the-art methods for temporal contrast synthesis. Readers were presented with the actual MRI acquisition alongside images generated by competing models, UNet \citep{unet} and TeNCA \citep{tenca}. They were tasked with selecting their preferred synthetic image based on a holistic assessment of tumor characterization accuracy, enhancement quality, and overall image fidelity. An optional free-text field was provided for readers to leave additional qualitative comments regarding their selection.

\subsubsection{Broader Implications}
Upon completion of the image-specific evaluations, readers participated in a final post-study survey designed to capture their broader perspectives on the current state and future utility of AI-based contrast synthesis. To assess the perceived maturity of the technology, readers were asked to evaluate how close the field is to solving the problem of contrast enhancement synthesis for clinical use, selecting from four levels of progress: very far (fundamental limitations remain), moderate progress (significant gaps remain), getting close (most cases are convincing), or nearly solved (clinically equivalent in most scenarios).

The remainder of the survey consisted of open-ended, free-text questions aimed at gathering qualitative insights to guide future technical development and clinical implementation. Readers were prompted to identify where current AI-based synthesis methods are most lacking and to specify which technical improvements they would prioritize (e.g., spatial resolution, temporal consistency, kinetic accuracy, or artifact reduction). Furthermore, the survey explored the readers' clinical vision for the technology, specifically asking for their thoughts on integrating virtual contrast MRI with standard mammography for breast cancer screening workflows, as well as its potential utility when combined with other unenhanced MRI sequences (such as diffusion-weighted imaging) for comprehensive diagnostic assessments. A final open-text field was provided to capture any remaining general feedback or observations regarding the study.

\subsection{Participants}

Four board-certified breast-imaging radiologists from three academic centers participated in the reader study. Sub-specialty experience ranged from 5 to 15 years (median 10.5, values 5, 8, 13, 15). Each reader evaluated all 40 cases across all four sections, yielding 2{,}927 analyzable responses (overall item-level missingness 12.9\%, predominantly the conditional and optional items).

\subsection{Statistical analysis}
Analyses were paired at the case$\times$reader level wherever the design
permitted. Categorical agreement between GT and synthetic characterization was
quantified with Cohen's $\kappa$ (unweighted for nominal tasks,
linear-weighted for ordinal tasks) and percent agreement; 95\% confidence
intervals were obtained by case-level bootstrap resampling (1{,}000 iterations).

Ordinal ratings (kinetic plausibility, image quality) were compared with the
Wilcoxon signed-rank test, with a \emph{signed} rank-biserial correlation $r$
(computed as $(T_{+}-T_{-})/(T_{+}+T_{-})$, negative values indicating lower
synthetic ratings) as the effect size, and Kendall's $\tau$ for
association. Inter-reader agreement on GT and on synthetic images was summarized
with Fleiss' $\kappa$ and bootstrap CIs.

The clinical-impact outcome (no change / minor / major) was modeled with a
proportional-odds ordinal logistic regression on standardized predictors
(perceived synthetic appearance, diagnostic complexity, image quality, reader
experience, and the three stratification axes); odds ratios with 95\% CIs are
reported. Method preference was tested globally with Cochran's $Q$ (the appropriate $K{=}3$
extension for related binary outcomes) and post-hoc with exact McNemar tests;
preference against chance used the binomial test. 
All families of tests were corrected for multiple comparisons within each analysis phase using the
Benjamini--Hochberg false-discovery-rate procedure at $q=0.05$; significance
($\alpha$) was set at 0.05.

\subsection{Additional results}

\subsubsection{Synthetic images add less variability than radiologists themselves}
\begin{figure}
    \centering
    \includegraphics[width=\linewidth]{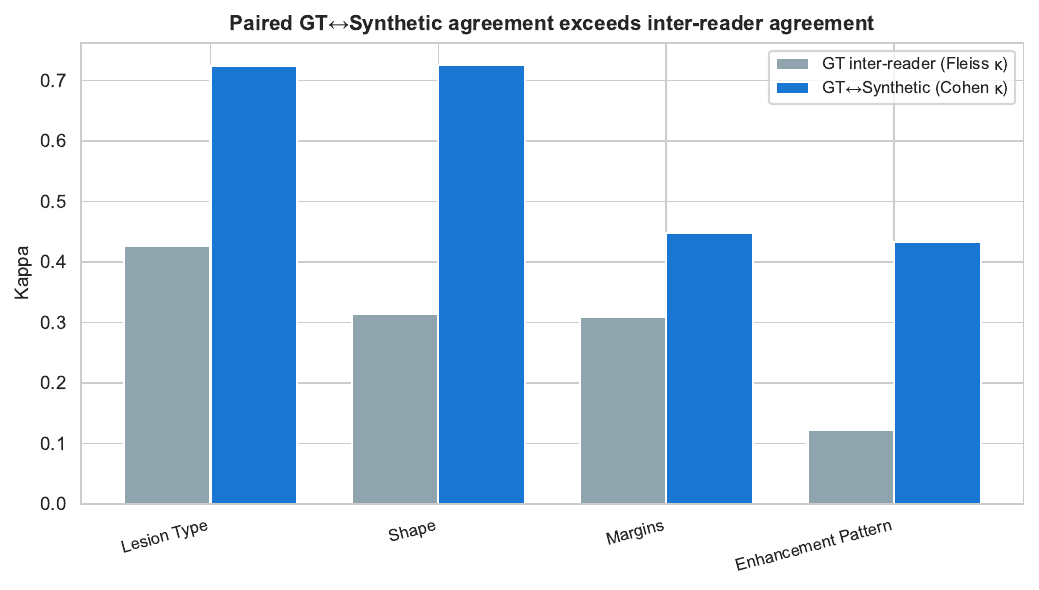}
    \caption{\textbf{Paired agreement exceeds inter-reader agreement.} For every categorical task, GT-to-synthetic Cohen's $\kappa$ (blue) exceeds GT inter-reader Fleiss' $\kappa$ (gray).}
    \label{fig:agree}
\end{figure}

For every categorical task, paired GT-to-synthetic Cohen's $\kappa$ exceeded
the GT \emph{inter-reader} Fleiss' $\kappa$ (Figure~\ref{fig:agree}). In other words, the discrepancy a synthetic image introduces relative to its own GT is smaller than the disagreement that already
exists between radiologists reading the same real images.

\subsubsection{Real images show have better image quality and kinetic plausibility} Extending the analysis in the main manuscript on this theme, Figure~\ref{appendix_fig:composite_results} show absolute number of cases binned in each category. Most real images are classified as having good kinetic curves, while for majority of synthetic curves are deemed plausible. Similarly, while the distribution of cases is similar in real and synthetic cases, the mean score is lower due to higher number of non-diagnostic and poor quality cases.

\subsubsection{Inter-reader agreement}
Inter-reader agreement was moderate for lesion type (Fleiss' $\kappa=0.43$ GT, 0.36 synthetic) and decreased for finer tasks, reaching near-zero for the two ordinal scales on both image types, indicating that kinetic-plausibility and quality ratings are inherently reader-subjective regardless of image origin.
\subsubsection{Failure mode}
\begin{figure}
    \centering
    \includegraphics[width=\linewidth]{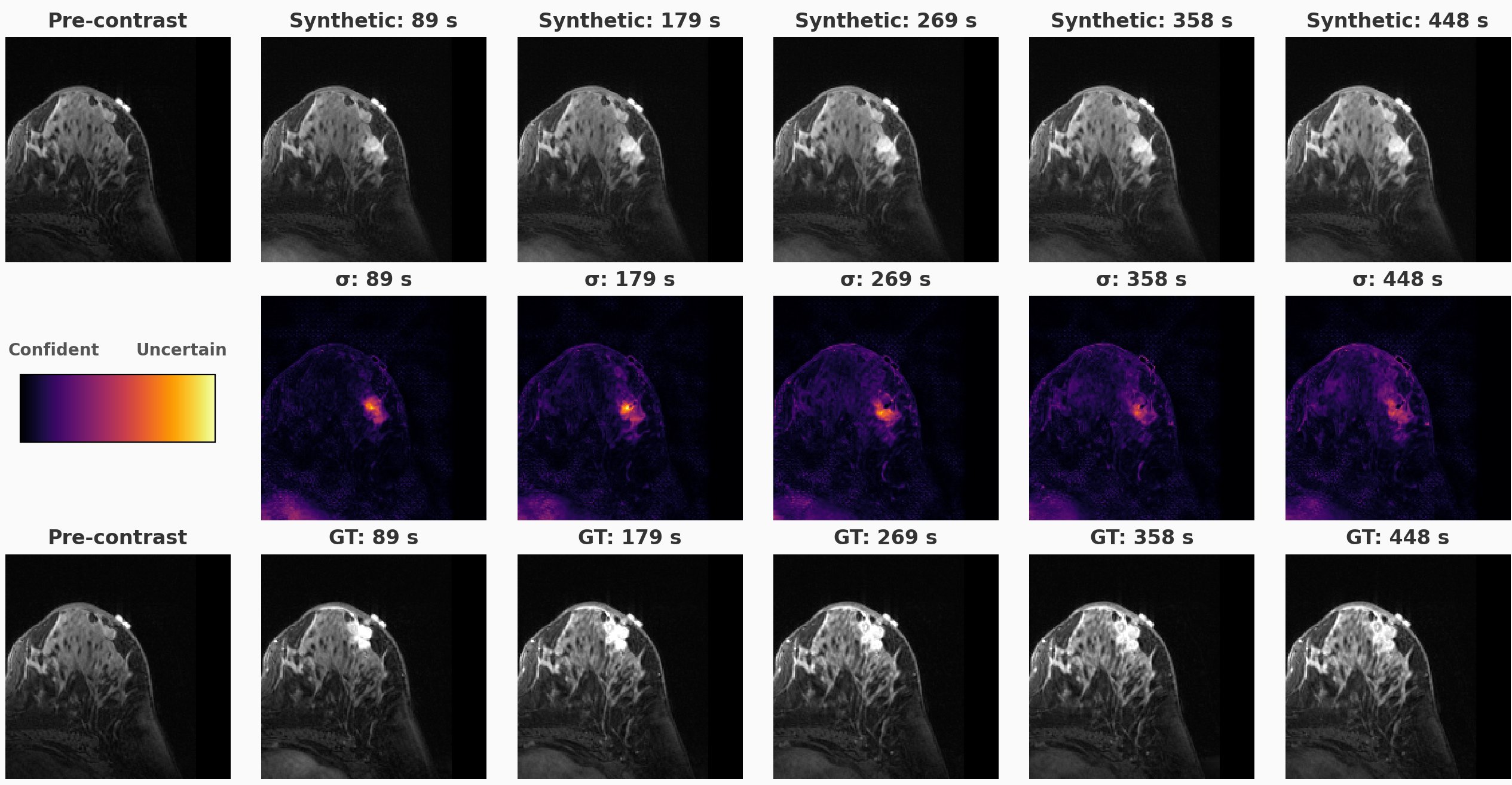}
    \caption{\textbf{Case demonstrating a major impact on clinical management.} Although the synthetic lesion is characterized identically to the ground truth, its incorrect spatial localization fundamentally alters the diagnostic outcome.}
    \label{fig:failure_clinical_study}
\end{figure}

Blinded uncertainty propagated to clinical concern: when a reader had answered
``cannot determine'' in the blinded phase, their later side-by-side impact rating
was markedly higher indicating worse impact (mean 1.48 vs.\ 0.85; 65\% vs.\ 24\% ``major''; Kendall's
$\tau$=0.25, $p<0.001$). Four cases drew unanimous (4/4) major-impact
verdicts. Three were explained by visible quality degradation, but one (case~31)
was rated indistinguishable from GT (quality $\Delta\approx0$) yet drew
unanimous major-impact verdicts because of spatial errors, specifically incorrect tumor
localization and extent, corroborated by three independent readers
(Case 31; Fig.~\ref{fig:failure_clinical_study}). This dissociation between perceptual realism and
spatial reliability is the study's cautionary finding: a synthetic image
can look entirely natural while placing or sizing disease incorrectly.

\section{MISCELLANEOUS}
\begin{figure}
    \centering
    \includegraphics[width=0.9\linewidth]{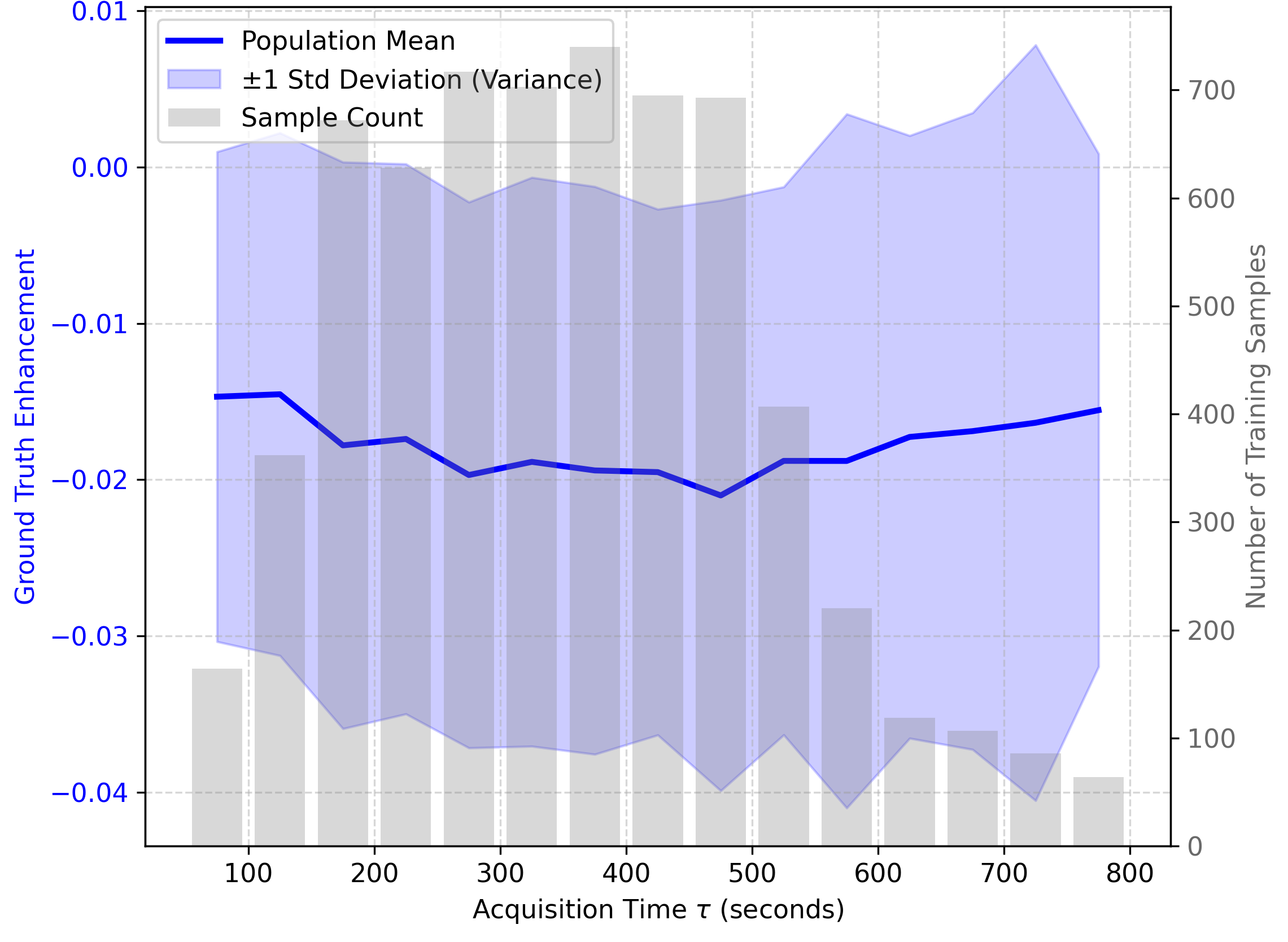}
    \caption{\textbf{Temporal distribution of training samples and ground truth contrast enhancement.} The overlaid histogram (grey bars, right axis) illustrates the number of available training samples across the acquisition time $\tau$, highlighting the long-tailed distribution with sparse representation in early ($<100$ s) and late ($>500\text{ s}$)  phases. The solid blue line and shaded region (left axis) denote the population mean and $\pm 1$ standard deviation of the ground truth enhancement trajectory, respectively.}
    \label{appendix_fig:data_distribution}
\end{figure}

\begin{table*}
\scriptsize
\centering
\caption{\textbf{Ablation Study.}    For a fraction of training samples, we replace the real post contrast objective with a synthetic sample where tau falls in a sparsely-populated region of the acquisition-time distribution. The post-contrast latent is linearly interpolated between the two nearest available phases (including pre at tau=0). This aims to give the model smooth, learnable supervisory signal across the entire    $\tau$ range, especially in regions where real training data is scarce. Using this sampling strategy, the model still produces contrast and shows improvement over baseline. However, it is not able to learn physiologically accurate temporal kinetics, with marked increase in peak timing error. Best results are highlighted in \textbf{bold}. MSE, DTW and DTW-ROI are reported in $10^{-2}$ scale.}
\label{appendix_tab:sampling}
\begin{tabular}{lccccccccccc}
\toprule
\rowcolor{pink!25}
\textbf{Method} & \textbf{MSE} $\downarrow$ & \textbf{PSNR} $\uparrow$ & \textbf{SSIM} $\uparrow$ & \textbf{LPIPS} $\downarrow$ & 
\textbf{FID - Dinov2} $\downarrow$ & \textbf{FRD} $\downarrow$ & \textbf{PTE} $\downarrow$ & \textbf{DTW} $\downarrow$ & \textbf{DTW-ROI} $\downarrow$\\
\midrule
\multicolumn{10}{l}{\textit{MAMA-MIA (Internal Validation)}} \\ \midrule

With Sampling & 0.88 (1.55) & 18.19 (4.29) & 0.71 (0.11) & 0.18 (0.05) & 161.65 & 5.29 & 60.79 (83.63) & 0.74 (0.60) & 3.76 (3.33) \\
Without Sampling & \textbf{0.84 (0.55)} & \textbf{21.61 (2.77)} & 0.71 (0.11) & \textbf{0.18} (0.05) & \textbf{154.56} & \textbf{4.98} & \textbf{44.57 (81.59)} & \textbf{0.68 (0.56)} & \textbf{3.76 (3.33)} \\ \midrule
\end{tabular}
\end{table*}

 \textbf{Training Data Distribution}: Figure~\ref{appendix_fig:data_distribution} are referenced in the main manuscript Section~IV to highlight lower sample sizes early and late in acquisition period. 

 \textbf{Sampling}: Table~\ref{appendix_tab:sampling} shows an ablation with and without using sampling strategy to increase representation in early ($<100s$) and late ($>500s$) acquisition period. This is referenced in main manuscript Section~IV. We employ a stochastic temporal latent augmentation strategy during training. For a given fraction of the batch, the method generates synthetic training targets by sampling new acquisition times ($\tau_{new}$) located within sparsely populated temporal region. Once $\tau_{new}$ is sampled, the algorithm identifies the two closest available real acquisition phases that bracket this time point. The corresponding images are encoded into the latent space, and a linear interpolation is performed between these two bracketing latents based on the relative temporal position of $\tau_{new}$. The original post-contrast training target is then replaced with this newly interpolated latent and its corresponding timestamp. This data-driven augmentation provides the model with a smooth, continuous supervisory signal across the entire temporal range, enforcing physically plausible kinetic transitions in low-density regions.

\bibliographystyle{cas-model2-names}
\bibliography{references}
\end{appendices}
\end{document}